\documentclass[11pt]{article} 
\usepackage{url}
\usepackage{smile}
\usepackage{graphicx} 
\usepackage{epstopdf}
\usepackage{wrapfig}
\usepackage[colorlinks, linkcolor=blue, anchorcolor=blue, citecolor=blue]{hyperref}
\usepackage[margin=1in]{geometry}
\usepackage[normalem]{ulem}
\usepackage[export]{adjustbox} 
\usepackage{mathtools, cuted}
\usepackage{enumerate}
\usepackage{enumitem}
\usepackage{microtype}
\usepackage{graphicx}
\usepackage{booktabs}
\usepackage{multirow} 
\usepackage{subcaption}
\usepackage{amsmath}
\usepackage{amssymb}
\usepackage{mathtools}
\usepackage{amsthm}
\usepackage{algorithm}
\usepackage{algorithmic}
\usepackage{wrapfig}

\usepackage[margin=1in]{geometry}
\usepackage[normalem]{ulem}
\usepackage[export]{adjustbox}
\usepackage{mathtools, cuted}
\usepackage{natbib}
\usepackage{enumerate}
\usepackage{enumitem}

\usepackage{kpfonts}
\DeclareMathAlphabet{\mathsf}{OT1}{cmss}{m}{n}

\SetMathAlphabet{\mathsf}{bold}{OT1}{cmss}{bx}{n}

\newtheorem*{theorem*}{Theorem}
\newcommand{\Wcal}{\mathcal{W}}
\newcommand{\Wzero}{W^{(0)}}
\newcommand{\wij}{w_{ij}}
\newcommand{\Wid}{W_{i*}}
\newcommand{\Wdi}{W_{*i}}

\newcommand{\Ibar}{\overline{I}}

\newcommand{\U}{U}
\newcommand{\Ut}{\U^{(t)}}
\newcommand{\Utp}{\U^{(t+1)}}
\newcommand{\V}{V}
\newcommand{\Vt}{\V^{(t)}}
\newcommand{\Vtp}{\V^{(t+1)}}
\newcommand{\St}{S^{(t)}}
\newcommand{\Stp}{S^{(t+1)}}
\newcommand{\Stil}{\tilde{S}}
\newcommand{\Stilt}{\Stil^{(t)}}
\newcommand{\Stdi}{S^{(t)}_{*i}}

\newcommand{\Stiltdi}{\Stilt_{*i}}
\newcommand{\Scal}{\mathcal{S}}
\newcommand{\Scalt}{\Scal^{(t)}}
\newcommand{\Sc}{\Gamma}
\newcommand{\Proj}{\mathcal{T}}

\newcommand{\OurAlg}{LoSparse}

\title{\bf LoSparse: Structured Compression of Large Language Models based on
Low-Rank and Sparse Approximation \footnote{Published as a conference paper in ICML 2023.}}

\author{Yixiao Li$^{**}$, Yifan Yu$^{**}$, Qingru Zhang, Chen Liang, \\ Pengcheng He, Weizhu Chen, Tuo Zhao \footnote{Li, Yu, Zhang, Liang and Zhao are affiliated with Georgia Tech. He and Chen are affiliated with Microsoft Azure. Correspondence to \url{yixiaoli@gatech.edu}, \url{yyu429@gatech.edu} and \url{tourzhao@gatech.edu}.}}

\newcommand{\commentout}[1]{}

\begin{document}

\maketitle
\def\thefootnote{**}\footnotetext{Equal contributions}

\begin{abstract}
Transformer models have achieved remarkable results in various natural language tasks, but they are often prohibitively large, requiring massive memories and computational resources. To reduce the size and complexity of these models, we propose {\OurAlg} (\textbf{Lo}w-Rank and \textbf{Sparse} approximation), a novel model compression technique that approximates a weight matrix by the sum of a low-rank matrix and a sparse matrix. Our method combines the advantages of both low-rank approximations and pruning, while avoiding their limitations. Low-rank approximation compresses the coherent and expressive parts in neurons, while pruning removes the incoherent and non-expressive parts in neurons. Pruning enhances the diversity of low-rank approximations, and low-rank approximation prevents pruning from losing too many expressive neurons. We evaluate our method on natural language understanding, question answering, and natural language generation tasks. We show that it significantly outperforms existing compression methods. Our code is publicly available at \url{https://github.com/yxli2123/LoSparse}

\end{abstract}

\section{Introduction}\label{sec:introduction}

Large transformer models have exhibited superior performance in various natural language tasks, such as natural language understanding, question answering, and natural language generation \citep{devlin2018bert, liu2019roberta, he2020deberta, radford2019language, brown2020language}. However, these models contain billions of parameters. 
For example, T5 \citep{radford2019language} consists of up to 11 billion parameters, and GPT-3 \citep{brown2020language} comprises up to 175 billion parameters. 
Their extreme sizes bring challenges in deploying the models to practical applications due to memory and computational requirements.  

To circumvent aforementioned challenges, model compression methods are widely applied to reduce model size at only a small expense of model performance. 
One common compression technique is pruning \citep{zhu2017prune, louizos2017learning}, which removes parameters according to their importance scores \citep{han2015learning, molchanov2016pruning, zhang2022platon}. Pruning methods can be divided into two categories: structured and unstructured pruning. In structured pruning \citep{mccarley2019structured,fan2019reducing,lagunas2021block}, 
weight matrices are pruned neuron/column-wise. This enables us to store pruned models by directly deleting neurons/columns in memory. As for unstructured pruning \citep{han2015learning, sanh2020movement}, however, 
weight matrices are pruned entry-wise, which makes it challenging to store and manipulate. For this reason, we focus on structured pruning. 
One popular structured pruning method is iterative pruning (ITP), which conducts training and pruning simultaneously. That is, after parameters are updated every iteration, it evaluates the importance score of each neuron. Neurons that have low importance scores are considered non-expressive and should be pruned. Beside the ITP, Movement pruning \citep{sanh2020movement} and CoFi \citep{xia2022structured} are also popular pruning methods.

\begin{figure}[htb!]
\centering
\begin{subfigure}{0.49\columnwidth}
    \centering
    \includegraphics[width=0.7\textwidth]{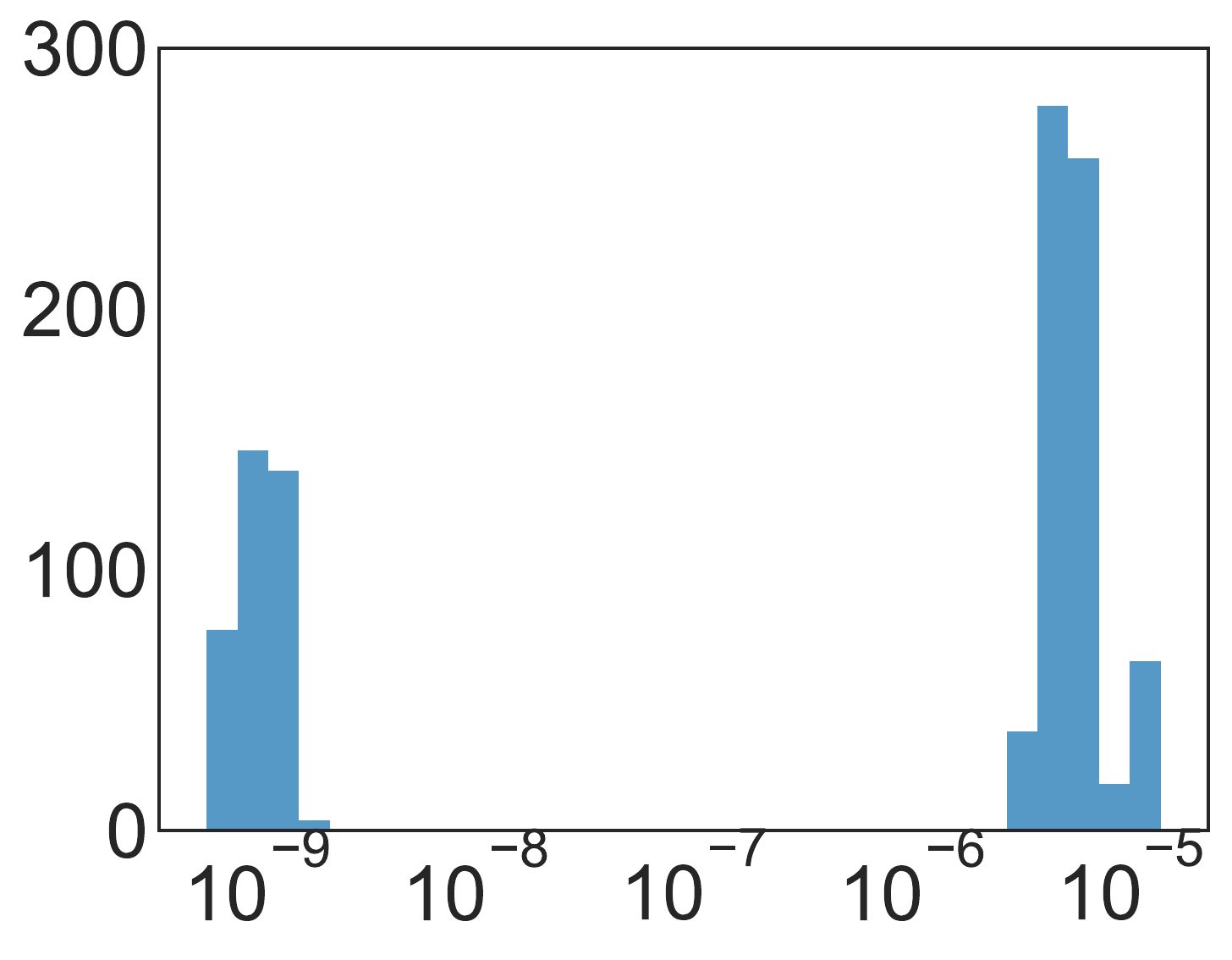}
    \caption{Query of decoder layer 9}
    \label{fig:ipt_hist_real1}
\end{subfigure}
\begin{subfigure}{0.49\columnwidth}
    \centering
    \includegraphics[width=0.7\textwidth]{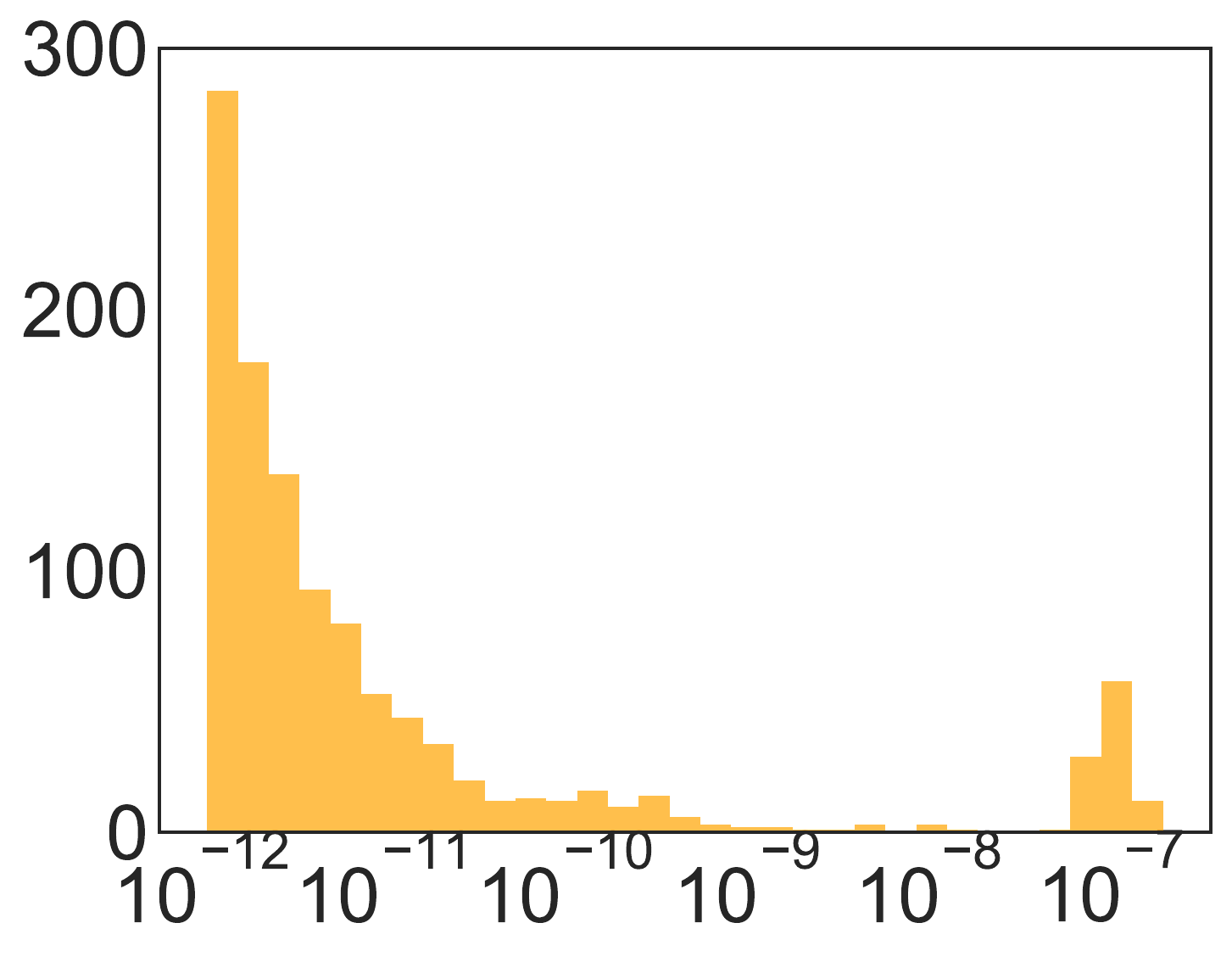}
    \caption{Ideal}
    \label{fig:ipt_hist_ideal}
\end{subfigure}
\vspace{-3mm}
\caption{Histogram of neuron importance scores. (a) The practical neuron importance scores of a linear layer when pruning BART-large on XSum. (d) The ideal histogram of the neuron importance scores where most of the neuron should be redundant, otherwise pruning is not the best choice.}
\label{fig:ipt_hist}
\vspace{-5mm}
\end{figure}

Unfortunately, pruning is not necessarily effective. It will inevitably remove expressive neurons given a high sparsity level. \citet{liang2021super} found heavy pruning hurts the performance severely, although light pruning can enhance the generalization of pre-trained language models. As an example, Figure~\ref{fig:ipt_hist} illustrates this phenomenon. Ideally (Figure \ref{fig:ipt_hist_ideal}), most of the neurons should be redundant and have low importance scores so that we can remove these neurons without hurting the performance too much. However, in reality (Figure \ref{fig:ipt_hist_real1}), even if a significant portion of neurons are non-expressive, the majority of neurons are still expressive and are likely to be pruned if the sparsity level is high.


Another popular compression technique is low-rank approximation \citep{hsu2022language, hajimolahoseinicompressing, tahaei2021kroneckerbert}, which is designed to compress the expressive neurons. It approximates a weight matrix by a low-rank matrix that is computed by singular value thresholding. 
Such a low-rank approximation is particularly effective to compress coherent parts in neurons. 
For example, the majority of neuron weights often share one coherent subspace that can be well approximated by singular vectors of top singular values. 
The low-rank approximation is inherently capable of extracting the common bases of this coherent subspace.


Matrices in transformer models, however, are often high-rank. 
It can hurt the model performance when merely applying the low-rank approximation to compress these matrices. 
This is because the diversity of neurons has been ignored. 
Although low-rank approximation extracts the common bases shared by neuron weights, 
it cannot accurately approximate their incoherent parts, 
which can be expressive and crucial to the model performance. We explain the reason more in Section~\ref{sec:low_rank_sparse}. 

To overcome the drawbacks of both pruning and low-rank approximations, we propose {\OurAlg} (Low-Rank and Sparse approximation), which approximates a weight matrix by the sum of a low-rank matrix and a sparse matrix. 
Such a composite approximation decouples the coherent parts from incoherent parts of neurons. 
It inherits the benefits of both low-rank and sparse approximation: 
the low-rank approximation aims to compress expressive bases of the coherent subspace shared by neurons while the sparse approximation focuses on removing non-expressive information in incoherent parts of neurons. 
In that sense, the low-rank approximation prevents the pruning from excessively removing expressive neurons while sparse approximation enhances the diversity of low-rank approximation. 



We draw inspiration from multi-task learning \citep{jalali2010dirty}, where linear models are used for multi-task regression. In out settings, every linear layer in transformer models can be naturally viewed as a linear multi-task model that learns different latent features. In that case, low-rank approximations are designed to store shared features across all coherent parts of neurons, and sparse approximations aim to learn distinct features from incoherent parts of neurons. Besides, previous work \citep{yu2017compressing, hawkins2021low, chen2021dsee} applied a similar method to Convolutional Neural Networks (CNN) and parameter-efficient fine-tuning, but we will discuss the limitation of their methods in Section \ref{sec:discussion}.



We conduct extensive experiments on natural language understanding, question answering, and natural language generation tasks to demonstrate the effectiveness and efficiency of {\OurAlg}. On the natural language understanding tasks in GLUE \citep{wang2018glue}, our method significantly outperforms existing pruning methods. For example, on the MNLI dataset, {\OurAlg} achieves more than 2.0\% higher accuracy than existing baseline methods. On the question answering tasks in SQuADv1.1 \citep{rajpurkar2016squad}, our method surpasses other pruning methods by 3.3 points in F1 score under the extreme low remaining ratio \footnote{The proportion of retained parameters.}. On the natural language generation tasks in XSum \citep{Narayan2018xsum}, our method exceeds the current methods by 2.99 points in Rouge-1 score. Moreover, our method is orthogonal to the current knowledge distillation methods, and could be readily integrated with them to improve the performance.

\section{Background}

We briefly review the transformer language models and pruning methods.

    \subsection{Transformer Models} 
    A typical transformer architecture comprises several sequential layers, where each layer contains two sub-layers: a multi-head self-attention (MHA) and a fully connected feed forward network (FFN). 
    Given the input $X \in \RR^{n\times d}$, 
    MHA computes the attention in parallel $h$ heads: 
    \begin{align*}
        & \mathrm{MHA}(X) = \mathrm{Concat}(\mathrm{head}_1, ..., \mathrm{head}_h) W_o, \\
        & \mathrm{head}_i = \mathrm{Softmax}({XW_{q_i} (XW_{k_i})^T}/{\sqrt{d_h}}) XW_{v_i},
    \end{align*}
    where $ W_{q_i}, W_{k_i}, W_{v_i} \in \RR^{d \times d_h}$ are query, key, and value projection matrices, $W_o \in \RR^{d\times d}$ is an output projection matrix, and $d_h $ is typically set as $d/h$. 
    FFN comprises two linear transformations and an activation: $ \mathrm{FFN}(X) = \sigma(X W_{f_1} + b_1) W_{f_2} + b_2, $
    where $W_{f_1} \in \RR^{d \times d_m} $, $ W_{f_2} \in \RR^{d_m \times d}$, and $\sigma(\cdot)$ is the activate function. A residual connection is used and followed by a layer normalization.     
    
    Generally, we denote all the matrix multiplication in a transformer model as
    \begin{align}
        y = Wx,
    \label{eq:y=wx}
    \end{align}
    where $W \in \RR^{d_1 \times d_2}$ denotes any weight matrix in the model.
    
    We further denote the parameter set consisting of all trainable weight matrix by $\Wcal = \{ W_m\}_{m=1}^{M}$. 
    Unless specified otherwise, we use $W$ to represent any weight matrix and $\Wzero$ is its pre-trained value. We use $i, j$ to index the entry of matrices and denote $\wij$ as $ij$-th entry of $W$, $\Wid$ as the $i$-th row of $W$, and $\Wdi$ as the $i$-th column of $W$.

    \subsection{Importance Scores for Pruning}
    Pruning methods zero out redundant parameters according to their importance estimation. Parameters with high importance scores are retrained for fine-tuning while the others with low importance are zeroed out. Popular importance metrics include magnitude \citep{han2015learning}, sensitivity \citep{sanh2020movement,molchanov2019importance} and uncertainty \citep{zhang2022platon}. 
    Sensitivity of parameters is essentially designed to approximate the change of training loss $\cL$ when a parameter is zeroed out. If the removal of a parameter causes a large variation on the loss, then the model is sensitive to this parameter and we should retain it. Specifically, for a weight $\wij$, its sensitivity score is defined by the gradient-weight product: 
    \begin{equation}\label{eq:sensitivity}
        I(\wij) = \left|\wij\cdot\nabla_{\wij} \cL \right|. 
    \end{equation}
    Note that the calculation of $I^{(t)}$ is conditioned on the sampled mini-batch at the $t$-th iteration. It can induce high uncertainty due to stochastic sampling. To reduce the variability in \eqref{eq:sensitivity}, \citet{zhang2022platon} propose to smooth $I$ by: 
    \begin{align}\label{eq:smooth_sensitivity}
        \Ibar^{(t)}(\wij) = \beta \Ibar^{(t-1)}(\wij) + (1-\beta)|\wij^{(t)}\nabla_{\wij}\cL^{(t)}|
    \end{align}
    using exponential moving of average.
    

    \subsection{Structured Pruning}
    As mentioned in Section~\ref{sec:introduction}, there are two types of pruning methods: unstructured and structured pruning. 
    Sensitivity in \eqref{eq:sensitivity}, however, targets on unstructured pruning. We extend it to structured pruning and introduce neuron importance scores. For a linear projection represented as a weight matrix $W\in \RR^{d_1\times d_2}$, we define the importance score of its $i$-th neuron $\Wdi$ as 
    \begin{equation}\label{eq:neuron-importance-score}
        \Sc(\Wdi) = \frac{1}{d_1} \sum_{j=1}^{d_1} \Ibar(w_{ji}). 
    \end{equation}
    We further define $\Sc(W) = [ \Sc(W_{*1}),...,\Sc(W_{*d_2}) ]^{\top} \in \RR^{d_2}$. 
    
    


\section{Method}\label{sec:method}
\begin{figure}[htb!]
    \begin{center}
    \centerline{\includegraphics[width=0.86\columnwidth]{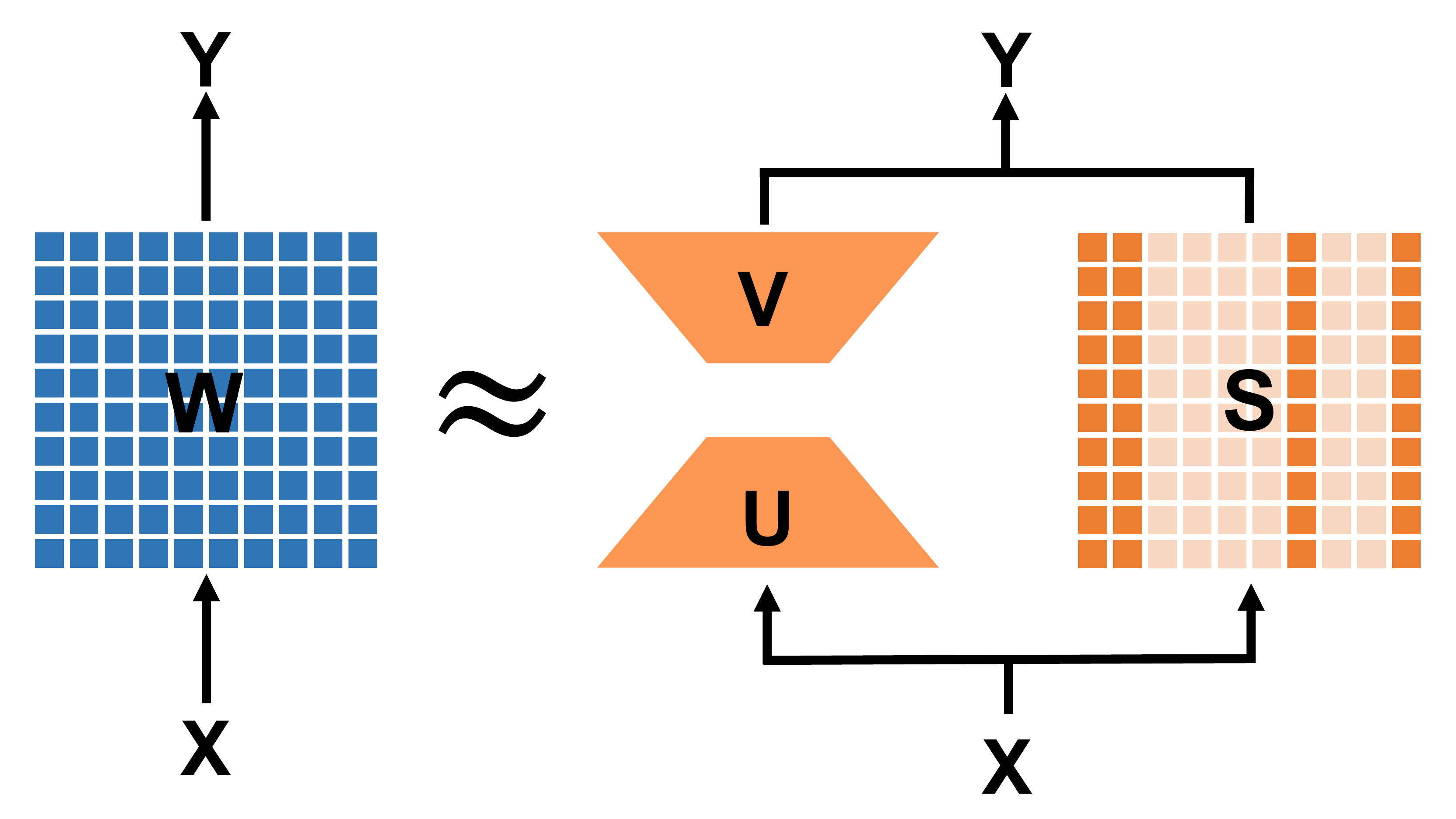}}
    \vspace{-3mm}
    \caption{Illustration of one linear projection in a transformer neural network. We use $UV +S$, a low-rank approximation plus a sparse matrix, to approximate the weight matrix $W$. $UV$ and $S$ indicate the coherent and incoherent parts of neurons in $W$ respectively. We conduct the forward pass of two terms in parallel.}
    \label{fig:illustration}
    \end{center}
    \vspace{-8mm}
\end{figure}

We propose a compression method for transformer models. Specifically, we approximate a weight matrix by the sum of a low-rank matrix and a sparse matrix (as illustrated by Figure \ref{fig:illustration}). 
The combination of these two approximations makes our compression method more efficient and stable.

\subsection{Approximation by Low-rank and Sparse Matrices}\label{sec:low_rank_sparse}
Given a weight matrix $W \in \RR^{d_1 \times d_2}$, a structured-pruned sparse matrix $S \in \RR^{d_1 \times d_2}$ is commonly applied to approximate $W$ for compression \citep{han2015learning,lagunas2021block}. The sparse matrix approximation, however, results in poor performance especially when the remaining ratio is low (See experiments in Section \ref{sec: experiment}). Therefore, we introduce a low-rank matrix to improve the approximation. Specifically, the weight matrix can be represented as
\begin{align}
    W = UV + S,
\label{eq:w=uv+s}
\end{align}
where the product of $U \in \RR^{d_1 \times r}$ and $V \in \RR^{r \times d_2}$ represents the low-rank matrix of rank $r$.

\begin{figure}[htb!]
\centering
\begin{subfigure}{0.47\columnwidth}
    \centering
    \includegraphics[width=0.7\textwidth]{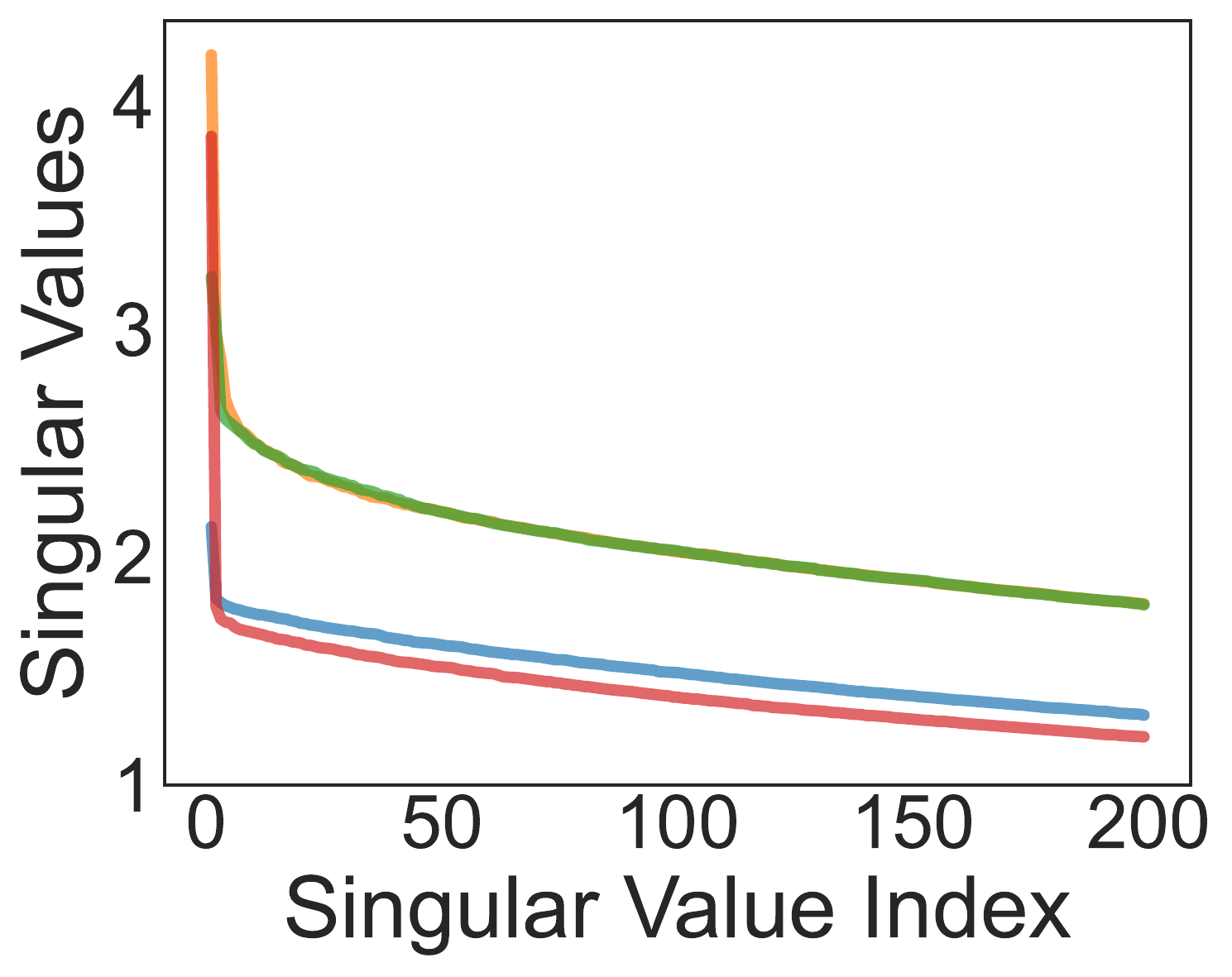}
    \caption{BART-large}
\end{subfigure}
\begin{subfigure}{0.49\columnwidth}
    \centering
    \includegraphics[width=0.7\textwidth]{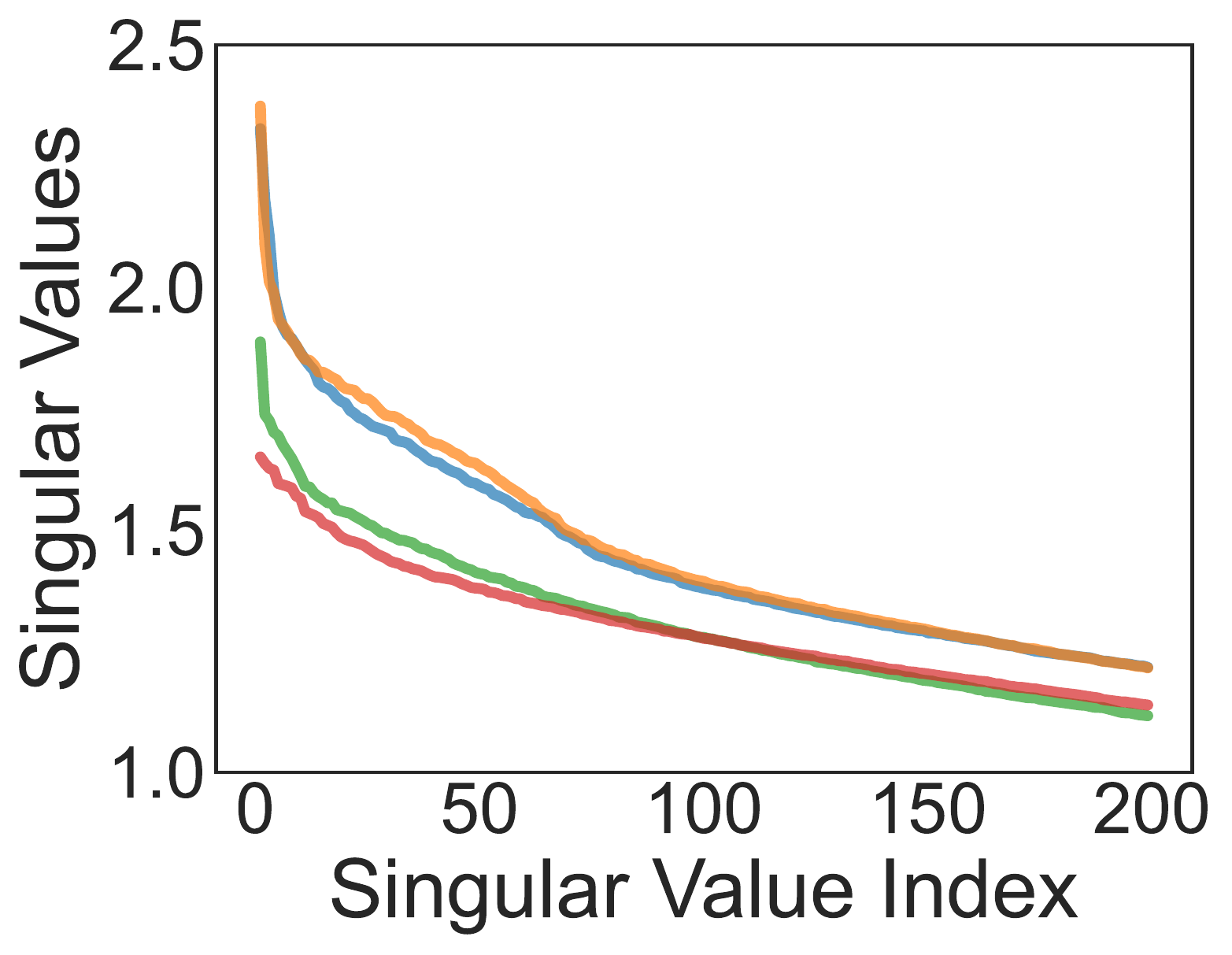}
    \caption{DeBERTaV3-large}
\end{subfigure}
\vspace{-3mm}
\caption{Singular values in language models. (a) Singular values of weight matrices of the 10\textsuperscript{th} decoder layer in BART-large; (b) Singular values of weight matrices of the 14\textsuperscript{th} encoder layer in DeBERTaV3-large.}
\label{fig:singular_value_distribution}
\vspace{-2mm}
\end{figure}

\textbf{Why low-rank matrices?} First, they can approximate the coherent parts of neurons effectively, even if the rank is small. 
As shown in Figure \ref{fig:singular_value_distribution}, we observe that the spectrum of weight matrices in language models drops rapidly at the beginning.  
This indicates neurons in a weight matrix share a common subspace, which can be viewed as the coherent parts of these neurons. In addition, the common subspace can be recovered by the singular vectors of top singular values.
Therefore, the coherent parts of neurons can be well approximated by the low-rank matrix computed by singular value thresholding.

\vspace{-1mm}

\begin{figure}[htb!]
\centering
\begin{subfigure}{0.49\columnwidth}
    \centering
    \includegraphics[width=0.7\textwidth]{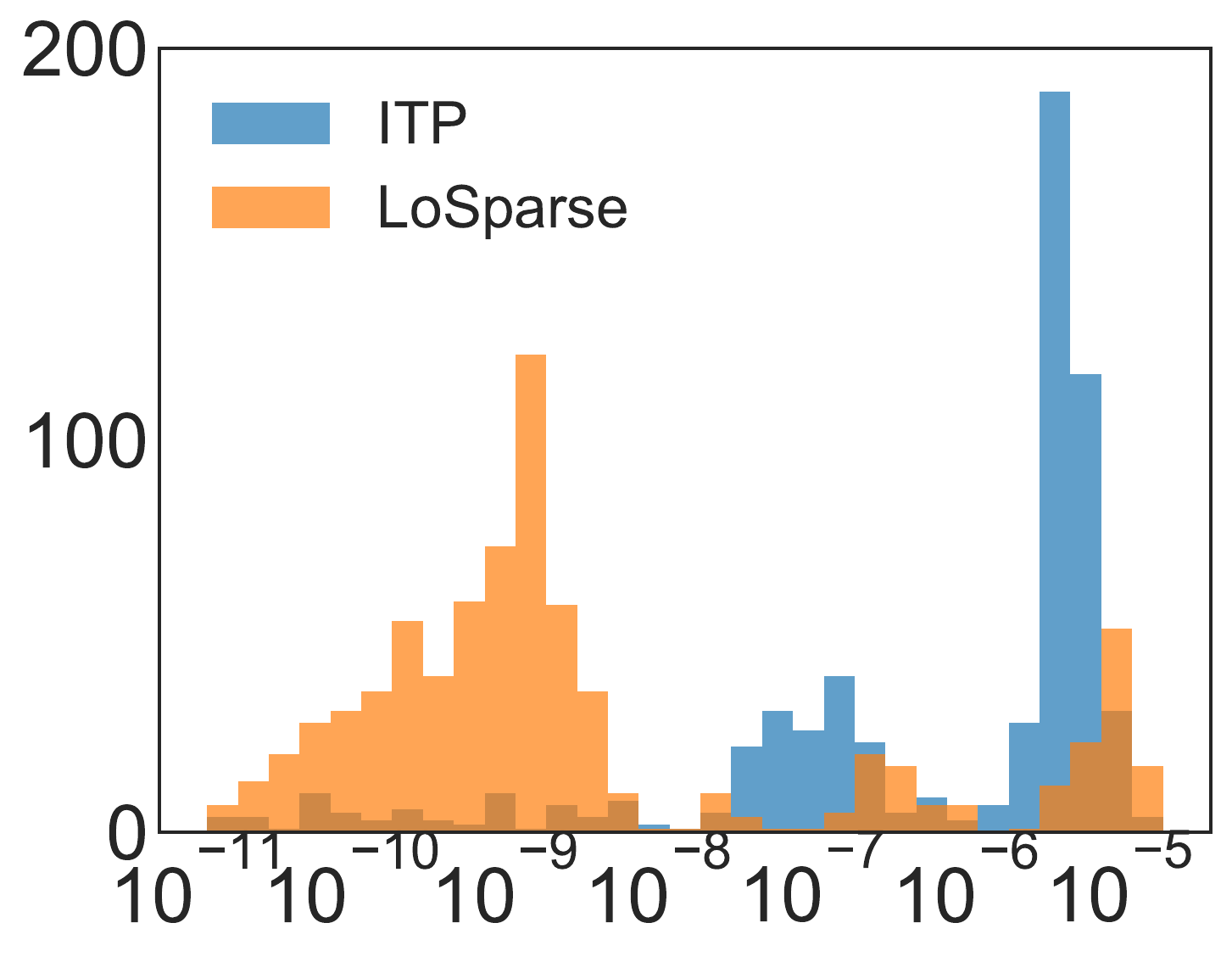}
    \caption{$W_{q}$ of Layer 3}
\end{subfigure}
\vspace{2mm}
\begin{subfigure}{0.49\columnwidth}
    \centering
    \includegraphics[width=0.7\textwidth]{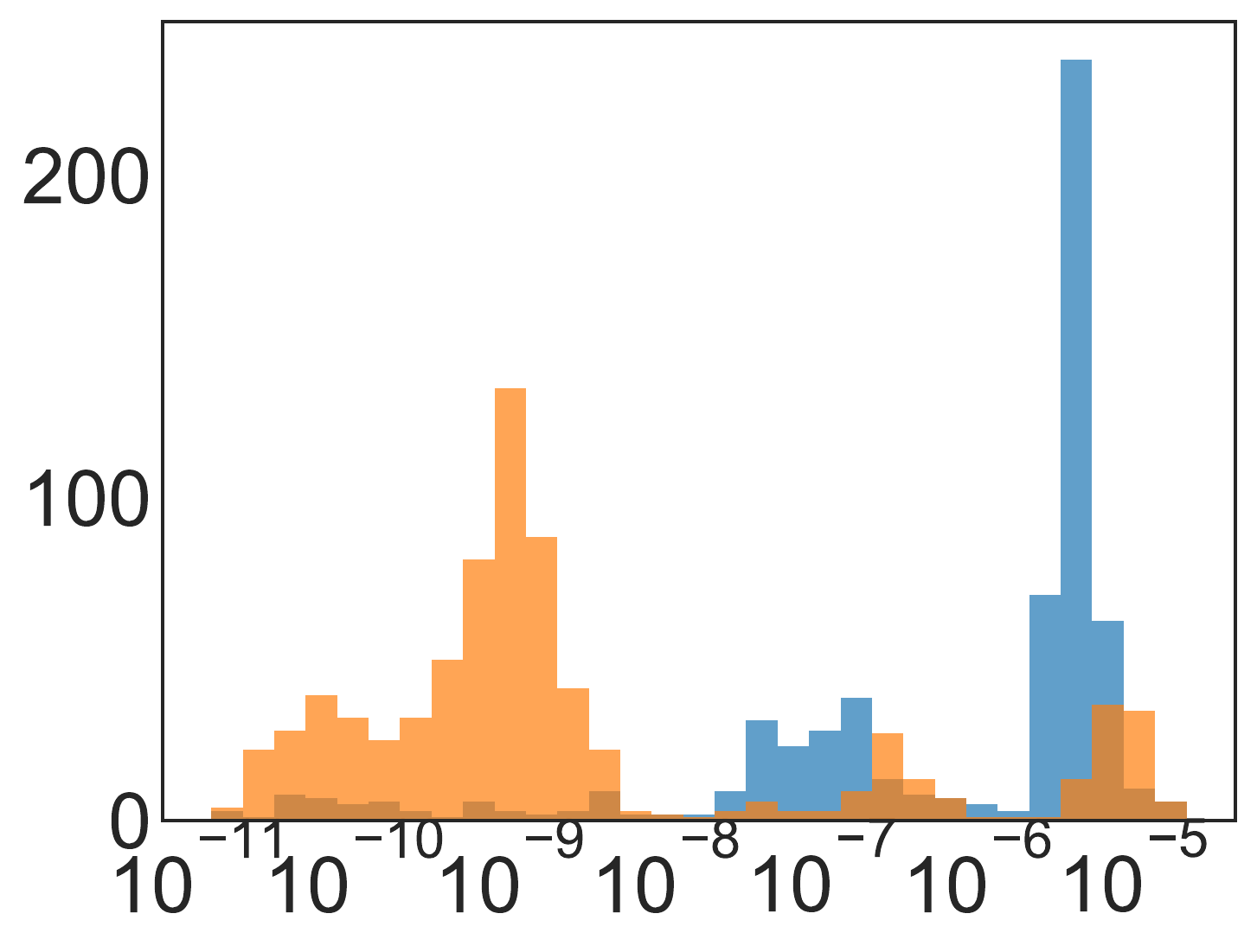}
    \caption{$W_{k}$ of Layer 3}
\end{subfigure}
\vspace{-5mm}
\caption{Neuron importance scores of selected linear projections when compressing DeBERTaV3-base on SST-2 with ITP (blue) and {\OurAlg} (orange). It shows {\OurAlg} successfully separates incoherent parts of neurons and make it easy to prune the non-expressive components. 
}
\vspace{-1mm}
\label{fig:shift}
\end{figure}

Second, the decoupling of low-rank and sparse matrices makes it easy to prune. 
The heavy-tailed spectrum in Figure~\ref{fig:singular_value_distribution} indicates each neuron $\Wdi$ spans their individual subspaces, which can represent the incoherent parts of these neurons. Since these subspaces are not shared, the incoherent parts cannot be captured by the low-rank approximation. Fortunately, the low-rank matrix is able to decouple the coherent parts from the incoherent parts of neurons. This enables us to approximate the remaining incoherent parts by adding a new matrix $S$ and then prune it to remove the non-expressive incoherent parts. 
As an example,  Figure~\ref{fig:shift} demonstrates that most of the incoherent parts have low importance scores after decoupling, motivating us to remove these redundant parameters.


\subsection{Algorithm}
    We then present our proposed algorithm. 
    Given a pre-trained weight matrix $ \Wzero $, we first initialize the low-rank matrix of rank $r$ based on the singular value decomposition (SVD) of $\Wzero$. Specifically, we choose 
    \begin{align}
        U^{(0)} &= [\sqrt{\sigma_1}u_1; \sqrt{\sigma_2}u_2; ... ;\sqrt{\sigma_r}u_r],
        \label{eq:left low rank matrix}\\ 
        V^{(0)} &= [\sqrt{\sigma_1}v_1; \sqrt{\sigma_2}v_2; ... ;\sqrt{\sigma_r}v_r]^\top,
        \label{eq:right low rank matrix}
    \end{align}
    where
    $u_1, u_2, ..., u_r \in \RR^{d_1}$ are left-singular vectors and  $v_1, v_2, ..., v_r \in \RR^{d_2}$ are right-singular vectors, with respect to the top $r$ singular values $\sigma_1 \ge \sigma_2 \ge ... \ge \sigma_r$ in the SVD of $W^{(0)}$. Then, we initialize $S^{(0)}$ by
    \begin{align}
        S^{(0)} = W^{(0)} - U^{(0)}V^{(0)}.
        \label{eq:initialize sparse matrix}
    \end{align}
    Notably, we replace the forward pass involving $W$ (e.g.~$Y = XW$) with \eqref{eq:y=uvx+sx} to improve computational efficiency:
    \begin{align*}
        Y = (XU)V + XS.
    \label{eq:y=uvx+sx}
    \end{align*}
    We apply such a decomposition to every weight matrix of the model and denote $\Scal = \{S_m\}_{m=1}^{M} $ as the set of all sparse matrices. After the initialization, we conduct the iterative structured pruning for $S$. Specifically, at $t$-th iteration, we first take a stochastic gradient decent step to update $\Ut, \Vt$, and $\St$. In particular, for $\St$,
    \begin{align*}
        \Stilt = \St - \alpha \nabla_{\St} \cL. 
    \end{align*}
    Then we evaluate the neuron importance scores of $\St$ based on \eqref{eq:neuron-importance-score}. Given the importance scores, $\Stilt$ is pruned following: 
    \begin{align}
        \Stp = \Proj(\Stilt, \Sc(\St))
    \end{align}
    with the $i$-th column of $\Proj(\Stilt, \Sc(\St))$ defined as
    \begin{align*}
        \Proj(\Stilt, \Sc(\St))_{*i} = \left\{\begin{array}{cc}
            \Stiltdi & \text{if $\Sc(\Stdi)$ in top $p_t$\%,} \\
            0 & \text{o.w.} \\
        \end{array}\right.
    \end{align*}
    Here we retain $\Stiltdi$ only if its importance score is in top $p_t$\% among all neurons in $\Scal^{(t)}$.   
    $p_t$ is the percentage of remaining neurons at $t$-th iteration. We gradually decay $p_t$ following a cubic schedule: 
    \begin{align*}\label{eq:compute threshold}
        p_{t} = 
        \begin{cases}
            1 & 0 \leq t<t_i, \\ 
            p_T+\left(1 -p_T\right)\left(1-\frac{t-t_i-t_f}{T-t_i-t_f}\right)^3 & t_i \leq t<T-t_f, \\ 
            p_T & \text {o.w.}
        \end{cases}
    \end{align*}
    where $T$ is the total training steps. $t_i$ is the number of initial warm-up steps. $t_f$ is the number of final fine-tuning steps. 
    Finally, we summarize our algorithm in Algorithm \ref{alg:main}.

    \begin{algorithm}[htb!]
    \caption{\OurAlg}
    \begin{algorithmic}[1]
    \STATE{\textbf{Input}: Pre-trained weights $\Wcal^{(0)}$; total training iterations $T$; the rank $r$; learning rate $\alpha$.}
    \FORALL{$W^{(0)} \in \Wcal^{(0)}$} 
    \STATE {Compute the SVD of $\Wzero$;}
    \STATE{Initialize $U^{(0)}$ and $V^{(0)}$ by \eqref{eq:left low rank matrix} and \eqref{eq:right low rank matrix};}
    \STATE{Initialize $S^{(0)} = W^{(0)} - U^{(0)}V^{(0)}$;}
    \STATE{Replace $\Wzero$ by $\U^{(0)}\V^{(0)} + S^{(0)}$;}
    \ENDFOR
    \FOR{$t = 1, ..., T$} 
        \STATE{Compute the gradient $\nabla \cL$;}
        \STATE{Compute $I^{(t)}$ for each parameter in $\Scalt$ by \eqref{eq:sensitivity};}
        \STATE {Compute $\Ibar^{(t)}$ for each parameter in $\Scalt$ by \eqref{eq:smooth_sensitivity};}
        \STATE {Compute $\Sc(\Stdi)$ for each $\Stdi$ in $\Scalt$ by \eqref{eq:neuron-importance-score};}
        \STATE {Update $\Stp = \Proj\left( \St-\alpha\nabla_{\St}\cL,  \Sc(\St) \right)$;}
        \STATE{Update $\Utp = \Ut - \alpha \nabla_{\Ut} \cL$;}
        \STATE{Update $\Vtp = \Vt - \alpha \nabla_{\Vt} \cL$;}
        
    \ENDFOR
    
    \STATE{\textbf{Output}: the compressed model. }\\
    \end{algorithmic}
    \label{alg:main}
    
\end{algorithm}

\begin{table*}[thb!]
\vspace{-1mm}
\caption{Results of pruned DeBERTaV3-base models on GLUE development set. Here {\it Ratio} is the proportion of total remaining weights. Results with {\it N.A.} indicate the model does not converge. The best results on each dataset are shown in bold.}
\label{tab:glue_datasets}
\begin{center}
\begin{small}
\begin{tabular}{l|l|cccccccc}
\toprule
\multirow{2}*{\bf Ratio} & \multirow{2}*{\bf Method} & {\bf MNLI} & {\bf RTE} & {\bf QNLI}  & {\bf MRPC} & {\bf QQP } & {\bf SST-2} & {\bf CoLA} & {\bf STS-B} \\  
~ & ~ & {m / mm} & {Acc} & {Acc} & {Acc / F1} & {Acc / F1} & {Acc} & {Mcc} & { P/S Corr} \\
\midrule 
{\bf 100\%} & {$\text{DeBERTaV3}_{\text{base}}$} & {90.5 / 90.6} & {82.0} & {94.0} & {89.5 /  93.3} & {92.4 / 89.8} & {95.3} & {69.2} & {91.6 / 91.1} \\
\midrule
\multirow{3}*{\bf 20\%} & {Movement} & \emph{N.A} & 61.2 & {86.0} & {79.2 / 85.0} &  \emph{N.A.} & {89.4} & \emph{N.A.}  & {84.3 / 84.3} \\
~ & {ITP} & {82.8 / 82.5} & \emph{N.A.} & {87.8} & {82.0 / 87.0} & {90.0 / 86.4} & {90.8} & 49.0 & {87.4 / 87.0}  \\
~ & {\OurAlg} & {\bf84.5 / 83.8 } & {\bf68.0} & {\bf88.6} & {\bf85.0 / 89.4} & {\bf90.6 / 87.2} & {\bf91.7} & {\bf50.0} & {\bf88.8 / 88.5} \\
\midrule
\multirow{3}*{\bf 15\%} & {Movement} & \emph{N.A.}  &  59.0  & {N.A} & {78.5 / 84.3} &  \emph{N.A.} & {89.0} & \emph{N.A.} & 83.9 / 83.9\\
~ & {ITP} & {81.7 / 81.3} & \emph{N.A.} & {85.4} & {80.5 / 86.3} & {89.1 / 85.2} & {89.3} & 45.8 & {86.8 / 86.3}  \\ 
~ & {\OurAlg} & {\bf83.3 / 82.9} & {\bf66.9} & {\bf87.6} & {\bf83.6 / 88.0} & {\bf90.3 / 87.0} & {\bf90.4} & {\bf46.8} & {\bf87.7 / 87.3} \\
\midrule
\multirow{3}*{\bf 10\%} & {Movement} &  \emph{N.A.} & \emph{N.A.} & {N.A} & {77.0 / 83.4} &  \emph{N.A.} & {88.0} & \emph{N.A.} & \emph{N.A.} \\
~ & {ITP} & {79.7 / 79.6} & \emph{N.A.} & {82.3} & {78.5 / 84.3} & {88.3 / 84.4} & {88.3} & 38.0 & {86.3 / 86.0}  \\
~ & {\OurAlg} & {\bf81.7 / 81.8} & {\bf66.0} & {\bf86.1} & {\bf 82.3 / 87.4} & {\bf89.5 / 86.0} & {\bf89.2} & {\bf40.0} & {\bf87.2 / 87.0} \\
\bottomrule
\end{tabular}
\end{small}
\end{center}
\vspace{-2mm}
\end{table*}

\section{Experiments}
\label{sec: experiment}
We evaluate our method on natural language understanding (NLU), question answering (QA), and natural language generation (NLG) tasks. We apply {\OurAlg} for compressing DeBERTaV3-base \citep{he2021debertav3}, BERT-base \citep{devlin2018bert}, and BART-large models \citep{lewis-etal-2020-bart}.

\textbf{Implementation Details.} Following the prior work \citep{louizos2017learning, sanh2020movement,zhang2022platon}, we compress all the backbone weight matrices, except LayerNorm and final prediction head. 
Our implementation is based on publicly available {\it Huggingface Transformers} code-base \citep{NEURIPS2019_9015}. All the experiments are conducted on NVIDIA V100 GPUs. 

\textbf{Baselines.} We compare {\OurAlg} with the following baseline methods: 
\begin{itemize}
    \vspace{-3mm}
    \item \textit{Full fine-tuning} is the most common approach for adapting pre-trained model to down-stream tasks. The model is initialized with pre-trained weights and all model parameters are updated through a stochastic gradient decent. 
    \vspace{-3mm}
    \item \textit{Movement pruning} is an effective pruning method \citep{sanh2020movement}. It multiplies a trainable mask to each neuron during the the training. When the mask is smaller than a threshold, the corresponding neuron is pruned.
    \vspace{-3mm}
    \item \textit{Iterative pruning (ITP)} removes neurons directly when their importance scores are lower than a hard threshold at each iteration \citep{molchanov2019importance}.
\end{itemize}

\subsection{Natural Language Understanding}
\textbf{Models and Datasets.} We evaluate the performance of {\OurAlg} when pruning DeBERTaV3-base models on the General Language Understanding Evaluation (GLUE) benchmark \citep{wang2018glue}. GLUE includes two single-sentence classification tasks: SST-2 \citep{socher-etal-2013-recursive} and CoLA \citep{warstadt-etal-2019-neural}, and three similarity and paraphrase tasks: MRPC \citep{dolan-brockett-2005-automatically}, STS-B \citep{cer-etal-2017-semeval}, and QQP. There are also four natural language inference tasks in GLUE: MNLI \citep{williams-etal-2018-broad}, QNLI \citep{rajpurkar-etal-2016-squad}, RTE \citep{Dagan2007ThePR,BarHaim2006TheSP,giampiccolo-etal-2007-third,bentivogli2009fifth}, and WNLI \citep{levesque2012winograd}. Following previous works, we exclude WNLI in the experiments.

\textbf{Implementation Details.} We select the learning rates from $\{1\times10^{-5},3\times10^{-5}, 5\times10^{-5}, 8\times10^{-5},9\times10^{-5}, 1\times10^{-4}\}$. We select the proportion of the parameters of all low-rank matrices over all pre-trained parameters from $\{1\%, 2\%, 3\%, 5\%\}$. We discuss the influence of different proportion later in Section \ref{sec:analysis}. More implementation details, such as the training epochs and batch sizes, are presented in the Appendix \ref{sec:app_nlu}.

\begin{table}[htb!]
\vspace{-3mm}
\caption{Results of pruned BERT-base models on some of GLUE development sets. Here {\it Ratio} is the proportion of total remaining weights. Results with {\it N.A.} indicate the model does not converge. The best results on each dataset are shown in bold.}
\vspace{1mm}
\label{tab:glue_datasets_bert}
\begin{center}
\begin{small}
\begin{tabular}{l|l|ccc}
\toprule
\multirow{2}*{\bf Ratio} & \multirow{2}*{\bf Method} & {\bf MNLI} & {\bf RTE} & {\bf QNLI}   \\  
~ & ~ & {m / mm} & {Acc} & {Acc} \\
\midrule 
{\bf 100\%} & {$\text{Bert}_{\text{base}}$} & {84.5 / 84.6} & {70.5} & {91.3}\\
\midrule
\multirow{3}*{\bf 20\%} & {Movement} & 77.0 / 76.9  & \emph{N.A.} & {84.7}\\
~ & {ITP} & {80.1 / 79.8} & 64.4 & {86.5}\\
~ & {\OurAlg} & {\bf80.4 / 80.3} & {\bf65.2} & {\bf86.9}  \\
\midrule
\multirow{3}*{\bf 15\%} & {Movement} & 76.1 / 76.5   &  \emph{N.A.}  & 83.9\\
~ & {ITP} & {79.1 / 79.0} & 63.2 & {85.0}   \\ 
~ & {\OurAlg} & {\bf79.4 / 79.2} & {\bf64.3} & {\bf85.9}  \\
\midrule
\multirow{3}*{\bf 10\%} & {Movement} & 73.6 / 74.1 & \emph{N.A.} & 82.2  \\
~ & {ITP} & {77.7 / 78.3} & 61.8 & {83.9}  \\
~ & {\OurAlg} & {\bf 78.3 / 77.8} & {\bf63.0} & {\bf84.8} \\
\bottomrule
\end{tabular}
\end{small}
\end{center}

\end{table}

\begin{table*}[t!]
\caption{Results with DeBERTaV3-base and BERT-base on SQuAD v1.1. Here {\it Ratio} is the proportion of remaining weights. The best results on each dataset are shown in bold.}
\label{tab:squad_experiemnts}
\begin{center}
\begin{tabular}{l|cccccc}
\toprule
{\bf Ratio} & {\bf 5\%} & {\bf 10\%} & {\bf 20\%} & {\bf 30\%} & {\bf 40\%} & {\bf 50\%} \\
\midrule
$\text{DeBERTaV3}_{\text{base}}$ & \multicolumn{6}{c}{87.7 / 93.5} \\
{- ITP}   & 65.2 / 76.1  & 70.9 / 80.3  & 75.0 / 83.9  & 78.2 / 86.2  & 80.5 / 87.5  & 81.5 / 89.6 \\
{- \OurAlg} & {\bf 69.3 / 79.1 } & {\bf 72.9 / 82.8 } & {\bf 76.8 / 85.8 } & {\bf 80.2 / 88.0} & {\bf 82.1 / 89.4} & {\bf 82.3 / 90.3}\\
\midrule
$\text{BERT}_{\text{base}}$ & \multicolumn{6}{c}{80.9 / 88.2} \\ 
{- Movement}  & \emph{N.A. }& { 51.4 / 64.6} & {63.3 / 74.5} &   {68.8 / 79.0} &  {73.0 / 82.4} &  {76.2 / 84.1}\\

{- ITP} & 54.0 / 67.3 & { 62.5 / 74.2} & { 66.8 / 78.0} & {72.3 / 82.4} & {74.5 / 84.2} & {\bf 76.0 / 85.1}\\
{- \OurAlg} & {\bf  57.6 / 70.6}  & {\bf 65.2 / 76.8}  & {\bf 69.7 / 80.4}  & {\bf 73.0 / 82.9 }  & {\bf  74.6 / 84.2}  & { 75.8 / 85.1}
\\
\bottomrule
\end{tabular}
\end{center}

\end{table*}

\textbf{Main Results.} We compare our method with the baseline methods under different remaining ratios. The results are shown in Table \ref{tab:glue_datasets}. 
We see that {\OurAlg} achieves better or on par performance compared with existing approaches on all the datasets of GLUE under all remaining ratios. For example, when the remaining ratio is 10\%, {\OurAlg} achieves 81.7\% accuracy on MNLI-m dataset, which surpasses the best-performing baseline (ITP) by 2\%. In addition to the superior performance, our method is more stable than the baselines (e.g.~ITP and Movement). 
This is because each weight matrix in {\OurAlg} at least maintains a low-rank matrix $S$ and always updates it along training horizon. It prevents the dramatic variation of weight matrices from nonzero to zero. 
By contrast, weight matrices are possibly pruned to zero by other iterative pruning methods. The expressive parts in these weight matrices can alternate between being pruned and updated and finally leads to divergence.

Table \ref{tab:glue_datasets_bert} summarizes the results of pruning BERT-base on MNLI, RTE, and QNLI. Similar to Table \ref{tab:glue_datasets}, our methods outperforms all baselines under all sparsity level for all three datasets. For example, when the remaining ratio is 20\%, {\OurAlg} achieves 65.2\% accuracy on RTE dataset, while ITP only achieves 64.4\% accuracy. We remark that {\OurAlg} is even more effective under high sparsity level. For instance,  given the 10\% remaining ratio, {\OurAlg} outperforms ITP by 0.6\% on MNLI-m dataset (78.3 v.s.~77.7), 1.2\% on RTE (63.0 v.s.~61.8), and 0.9\% on QNLI (84.8 v.s.~83.9).

\subsection{Question Answering}
\textbf{Models and Datasets.} We evaluate the performance of our method on the question-answering task (SQuADv1.1, \citet{rajpurkar-etal-2016-squad}). In the SQuADv1.1, question answering is treated as a sequence labeling problem, where we predict the probability of each token being the start and end of the answer span. We compress DeBERTaV3-base and BERT-base on the SQuADv1.1.

\textbf{Implementation Details.}  We compress all the backbone weight matrices in DeBERTaV3-base model and BERT-base except layernorm and final classification head. We use learning rates from  $\{1\times10^{-5},3\times10^{-5}, 5\times10^{-5}, 8\times10^{-5}\}$ and pick the learning rate that performs the best. We also select the proportion of parameters of low-rank approximation from $\{1\%, 2\%, 3\%, 5\%\}$. We choose AdamW as the optimizer and set the batch size as 16. Please refer to the Appendix \ref{sec:app_squad} for more details. 

\textbf{Main Results.} We compare our method with the baseline methods under different sparsity levels. The experimental results are shown in Table \ref{tab:squad_experiemnts}. We can see that {\OurAlg} consistently surpasses baseline methods under all remaining ratios in terms of the two evaluation metrics: exact match (EM) and F1. Similar to our result in GLUE tasks, our method is especially effective with low remaining ratios. For example, {\OurAlg} outperforms ITP by 3.0\% in terms of F1 if removing 95\% of parameters. Even for high remaining levels, our method still achieves considerable performance gain. For example, {\OurAlg} outperforms ITP by 1.9\% in terms of F1 if removing 60\% of parameters. 

Table \ref{tab:squad_experiemnts} also summarizes pruning BERT-base with different methods on SQuADv1.1. Our method achieves substantial improvements compared to all baseline methods. For example, our method outperforms the best baseline ITP by 2.6\% on F1 given 10\% remaining ratio. We remark that ITP is also effective under low sparsity levels. For example, ITP achieves the best result over {\OurAlg} and movement pruning given the remaining ratio as 50\%. Our method, however, still behaves on par with ITP with high remaining ratios: both ITP and {\OurAlg} achieve 84.2 on F1 under 40\% remaining ratios.

\subsection{Natural Language Generation}
\textbf{Models and Datasets.} In natural language generation (NLG) tasks, we compress BART-large model \citep{lewis-etal-2020-bart} to compare {\OurAlg} with baseline methods. We evaluate the performance on the XSum \citep{Narayan2018xsum} and CNN/DailyMail\citep{hermann2015teaching} datasets.

\textbf{Implementation Details.} We apply our method to all weight matrices of both encoder and decoder layers. We report ROUGE 1/2/L scores, which are the metrics for summarization tasks \citep{lin-2004-rouge}. Given a fixed total remaining ratio, we try different allocations between the low-rank matrices and the sparse matrices. The best allocation of sparse matrices is 10\%. 
We choose the learning rate from $\{6\times10^{-6},1\times10^{-5}, 2\times10^{-5}, 4\times10^{-5}, 6\times10^{-5}, 1\times10^{-4}\}$. The training epochs and batch sizes are set to 12 and 32 respectively. The beam search length is 8. Please see Appendix \ref{sec:app_nlg} for the detailed configuration.

\begin{table}[htb!]
\caption{Results with BART-large on XSum. Here \textit{Ratio} is the proportion of remaining weights. We report R-1/2/L. The best results on each dataset are shown in \textbf{bold}. \textit{Lead-3} means choosing the first 3 sentences as the summarization.}
\label{tab:bart_summerization}

\begin{center}
\begin{small}
\begin{tabular}{c|c|cc}
\toprule
{\bf Ratio} & {\bf Method} & {\bf XSum} & {\bf CNN/DailyMail}  \\  
\midrule 
{-} & {$\text{Lead-3}$} & {16.30/1.60/11.95} & {40.42/17.62/36.67}  \\
{\bf 100\%} & {$\text{BART}_\text{large}$} & {45.14/22.27/37.25} &{44.16/21.28/40.90}  \\
\midrule
\multirow{2}*{\bf 50\%} & {ITP} & {38.42/16.32/31.43}  & {40.76/18.30/37.65} \\
~ & {\OurAlg} & {\bf39.18/\bf16.91/\bf31.62}  & {\bf41.54/\bf19.04/\bf38.58} \\
\midrule
\multirow{2}*{\bf 40\%} & {ITP} & {36.71/14.96/29.86}  & {40.52/18.10/37.31}\\ 
~ & {\OurAlg} & {\bf38.30/\bf16.02/\bf30.72} & {\bf41.42/\bf19.00/\bf38.47} \\
\midrule
\multirow{2}*{\bf 30\%} & {ITP} & {34.42/13.15/27.99}  & {40.35/17.98/37.15} \\
~ & {\OurAlg} & {\bf37.41/\bf15.42/\bf30.02}  & {\bf41.21/\bf18.84/\bf38.21} \\
\bottomrule
\end{tabular}
\end{small}
\end{center}

\end{table}

\textbf{Main Results.} We compare {\OurAlg} with baseline methods under 30\%, 40\%, and 50\% remaining ratios. We do not report results on lower remaining ratios because the baseline methods fail to surpass the Lead-3 baseline. Table \ref{tab:bart_summerization} summarizes experiment results on the XSum and CNN/DailyMail test datasets. Note that {\OurAlg} consistently surpasses the baseline methods under all remaining ratios in terms of ROUGE scores. For instance, {\OurAlg} outperforms ITP on XSum dataset by 2.99 in terms of ROUGE-1 score. We also remark that {\OurAlg} is more efficient under extremely low remaining ratios. For example, the gain on ROUGE-1 increases from 0.76 to 2.99 when the ratio drops from 50\% to 30\%. Note that {\OurAlg} is particularly effective on more difficult summarization tasks. For example, XSum is more abstract and hence more difficult than CNN/DailyMail, and {\OurAlg} yields 2.99 gain on XSum compared to 0.86 on CNN/DailyMail.

\begin{figure*}[thb!]
\centering
\begin{subfigure}{0.32\textwidth}
    \centering
    \includegraphics[width=0.9\textwidth]{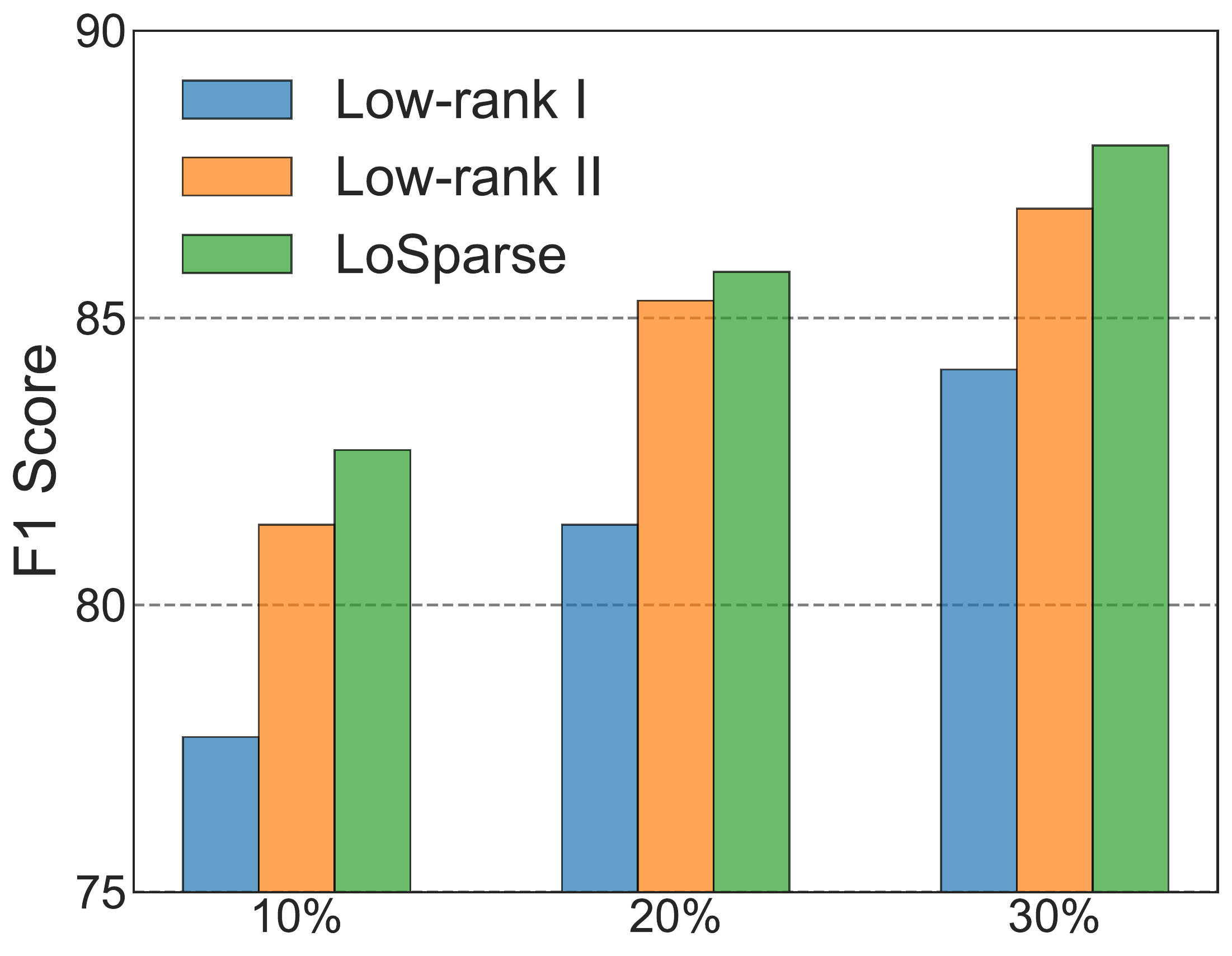}
    \vspace{-1mm}
    \caption{SQuAD v1.1}
\end{subfigure}
\begin{subfigure}{0.32\textwidth}
    \centering
    \includegraphics[width=0.9\textwidth]{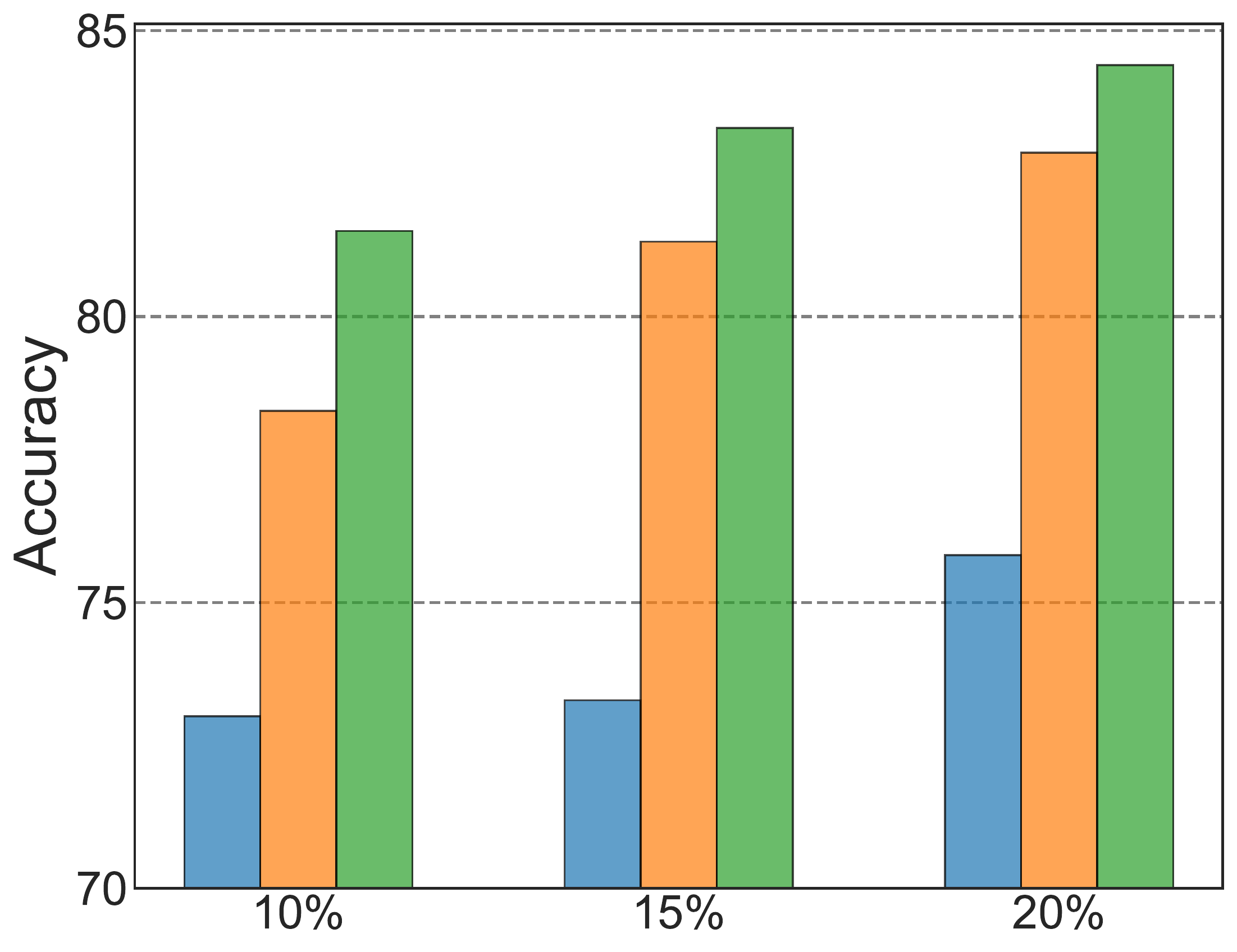}
    \vspace{-1mm}
    \caption{MNLI-m}
\end{subfigure}
\begin{subfigure}{0.32\textwidth}
    \centering
    \includegraphics[width=0.9\textwidth]{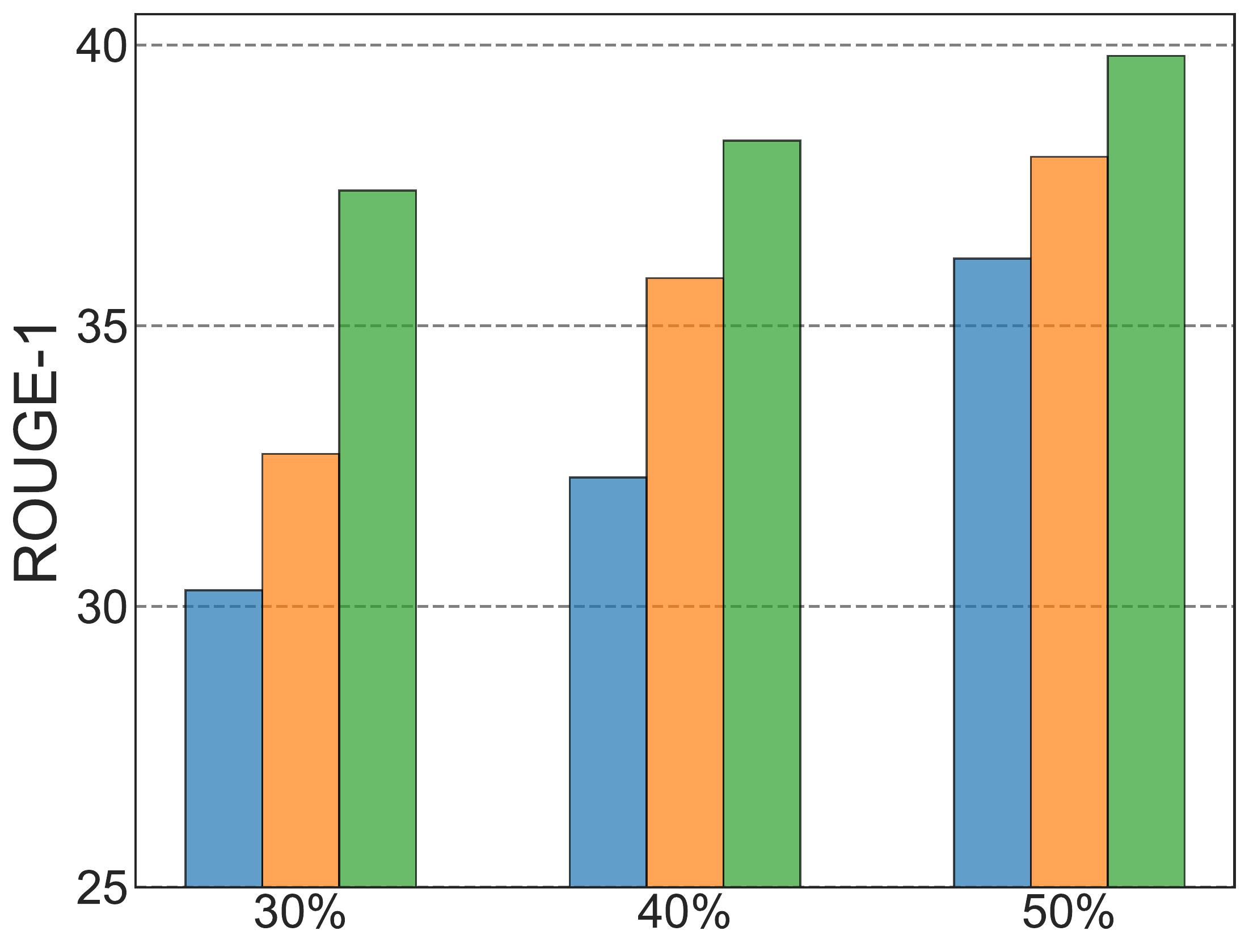}
    \vspace{-1mm}
    \caption{XSum}
\end{subfigure}

\caption{Comparison between {\OurAlg} and two variants of low-rank approximation on different tasks. The $x$-axis represents the remaining ratios. {\OurAlg} outperforms all other low-rank approximation variants. It indicates adding sparse approximation can promote the performance low-rank approximation.}
\label{fig:prune0}
\end{figure*}

\subsection{Analysis}
\label{sec:analysis}

\textbf{Effectiveness of Sparse Approximations.} We experiment the model compression without sparse approximation to study its effectiveness. Specifically, we compare {\OurAlg} with two variants: (i) we discard the sparse matrices and only fine-tune the low-rank matrices $\U\V$ (Low-rank I);  (ii) we follow the initialization as \eqref{eq:initialize sparse matrix} but gradually prune the initialized $S$ into zero (Low-rank II). Figure \ref{fig:prune0} summarizes the performance of these two variants on MNLI, SQuAD, and XSum. The results show that our method outperforms two low-rank variants, which verifies the effectiveness of the sparse approximation. Moreover, we find Low-rank II is better than Low-rank I. We discuss it in Section \ref{sec:discussion}.

\begin{figure*}[htb!]
\centering
\begin{subfigure}{0.32\textwidth}
    \centering
    \includegraphics[width=1.0\textwidth]{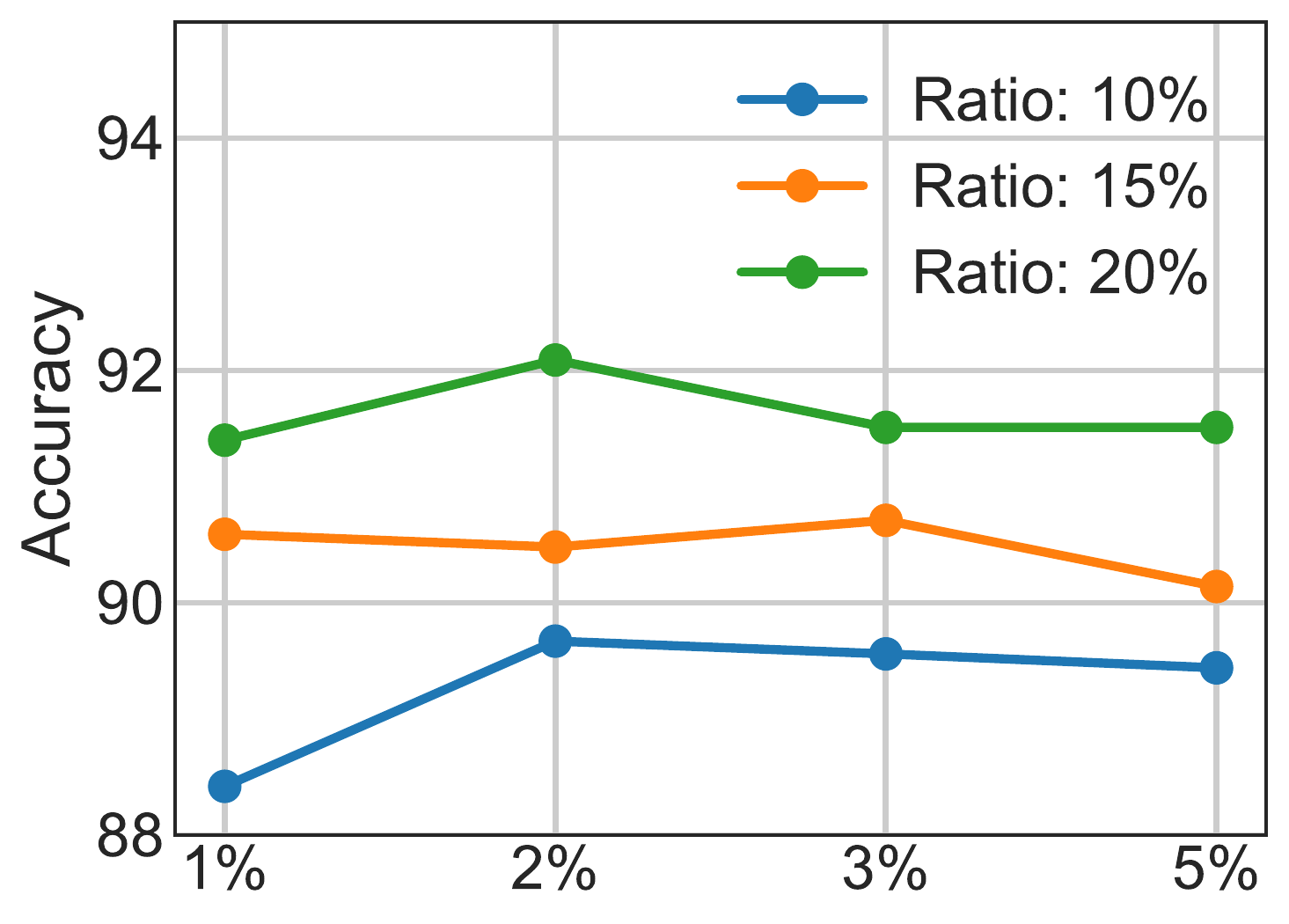}
    \vspace{-5mm}
    \caption{SST-2}
\end{subfigure}
\begin{subfigure}{0.32\textwidth}
    \centering
    \includegraphics[width=1.0\textwidth]{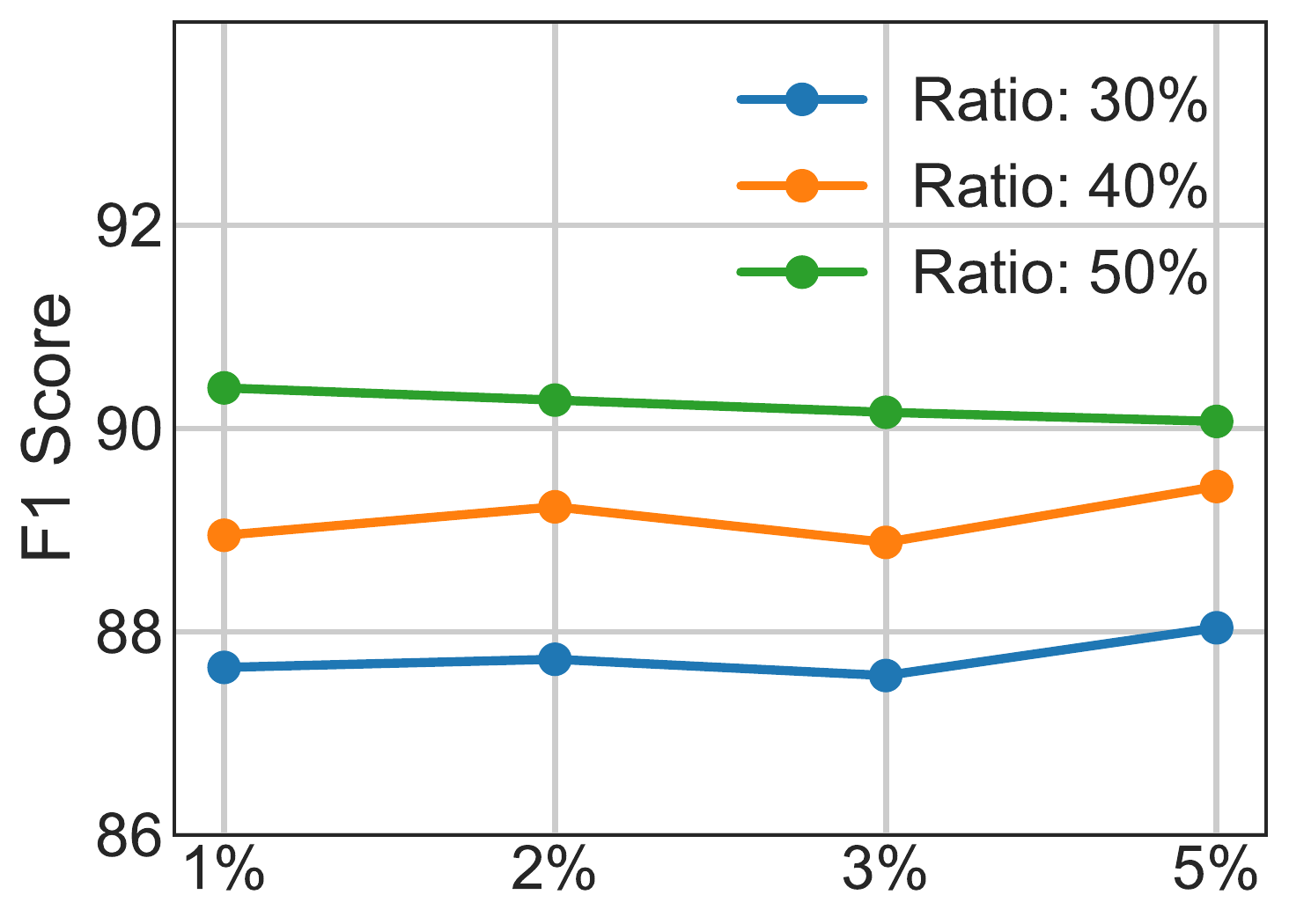}
    \vspace{-5mm}
    \caption{SQuAD v1.1}
\end{subfigure}
\begin{subfigure}{0.32\textwidth}
    \centering
    \includegraphics[width=1.0\textwidth]{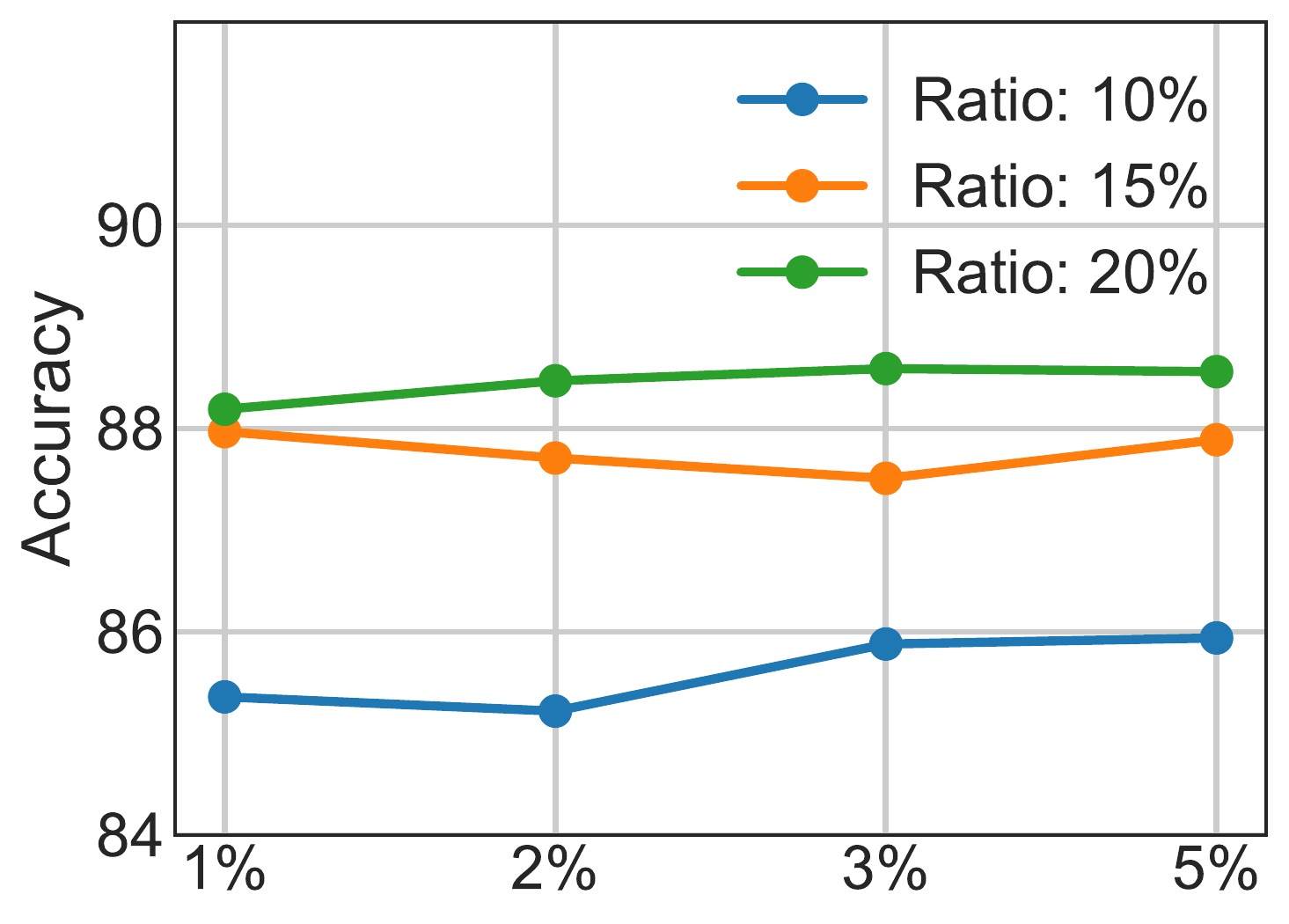}
    \vspace{-5mm}
    \caption{QNLI}
\end{subfigure}

\caption{Results about sparsity allocations. The $x$-axis represents the proportion of low-rank matrices over total pre-trained weights. The performance stays stable as changing the low-rank ratio. It suggests that our method is not sensitive to low-rank ratio.}
\label{fig:allocation}
\end{figure*}

\textbf{Sparsity Allocation.} We study how low-rank and sparse approximations cooperate with each other. Specifically, given a fixed remaining ratio, we change the proportion of low-rank matrices and accordingly the ratio of sparse matrices. Figure \ref{fig:allocation} summarizes the result under different allocations. We see low-rank and sparse approximations exhibit the nearly equal contribution to the performance on NLU tasks as the performance stays stable when changing the allocation. 

\subsection{Combination with Knowledge Distillation}
Knowledge distillation is a popular technique to improve the performance of small models \citep{romero2014fitnets, hinton2015distilling}. In knowledge distillation, the small model (student) is trained to mimic the output of a larger fine-tuned model (teacher) such that the performance of the small model can be improved. 

We remark that compression methods are complementary to knowledge distillation. We show it by integrating knowledge distillation into {\OurAlg} and other pruning methods. Specifically, we choose a DeBERTaV3-base model that is fine-tuned on specific tasks as the teacher model and a compressed DeBERTaV3-base model as the student model. Then we conduct layer-wise distillation for them. Please see Appendix \ref{sec:app_KD} for more training details. Table \ref{tab:glue_distillation} shows the results. 
We find that distillation can further improve the performance of {\OurAlg} and other compression methods. It shows that compression and knowledge distillation are complementary to each other. 
Besides, {\OurAlg} still achieves better performance than ITP when integrated with distillation. It demonstrates the effectiveness of {\OurAlg} in the setting of distillation. 

\begin{table}[htb!]

\caption{Results of distilling the fined-tuned DeBERTaV3-base to the compressed DeBERTaV3-base on MNLI, SQuAD, SST-2, and RTE. We compress the model to 20\% of its original size by {\OurAlg} and ITP, and then conduct layer-wise distillation.}
\label{tab:glue_distillation}
\begin{center}
\begin{tabular}{l|cccc}
\toprule
{\bf Method} & {\bf MNLI} & {\bf SQuAD} & {\bf SST-2} & {\bf RTE} \\ \midrule
$\text{DeBERTa}_{\text{base}}$ & 90.5 &87.6 / 93.5&96.1&82.0 \\ \midrule
{ITP}   & 85.9  & 77.9 / 87.2  & 92.0  & 58.1  \\
{\OurAlg}  & {\bf 84.6 / 84.7}  & {\bf 81.0 / 89.8} & {\bf 93.2}  & {\bf 71.1} \\
\bottomrule
\end{tabular}
\end{center}

\end{table}

We conduct further investigation on the performance of combining {\OurAlg} and knowledge distillation on BERT-base models. Similarily, we choose a BERT-base model  that is fine-tuned on specific tasks as the teacher model and a compressed BERT-base model as the student model. We find out combining {\OurAlg} and knowledge distillation can achieve a comparable or even better performance than popular compression method, such as PKD \citep{Sun2019PatientKD}, MixKD \citep{Liang2020MixKDTE}, CoDIR \citep{Sun2020ContrastiveDO}, BERT-of-Theseus \citep{Xu2020BERTofTheseusCB}, Low-Rank BERT Feature Distillation + KD \citep{Noach2020CompressingPL}, Path2: BERT+FWSVD\citep{Hsu2022LanguageMC}.
Please see Appendix \ref{sec:app_KD} for more training details.

Table \ref{tab:glue_distillation_bert} shows the comparison of the aforementioned methods. From the table , we can see that {\OurAlg} combined with knowledge distillation can achieve an on-par or better result than most distillation methods, such as PKD, under different remaining ratio. Our method also excels the combination of low-rank approximation methods and knowledge distillation such as Low-Rank BERT Feature Distillation + KD.





\begin{table*}[htb!]

\caption{Results of distilling the fined-tuned BERT-base to the compressed BERT-base on MNLI, RTE, QNLI, and SST-2. We compress the model to 25\% and 50\% of its original size by {\OurAlg}, and then conduct layer-wise distillation.}
\label{tab:glue_distillation_bert}
\begin{center}

\begin{tabular}{l|l|cccc}
\toprule
{\bf Ratio} & {\bf Method} & {\bf MNLI} & {\bf RTE} & {\bf QNLI} & {\bf SST-2}
\\ \midrule

50\% & PKD & 81.5 / 81.0 & 65.5 & 89.0 & 92.0 \\
&MixKD & 82.2 / 81.2 & 68.3 & 88.2 & 92.5 \\
&CoDIR-Fine & 83.6 / 82.8 & 65.6 & 90.4 & {\bf93.6 }\\
&  BERT-of-Theseus & 82.3 / N.A. & 68.2  & 89.5  & 91.5  \\
& Low Rank BERT Feature Distillation + KD & 84.8 / 83.7 & 71.1 & 91.4 & 92.4 \\
&  Path2: BERT+FWSVD & 83.0 / N.A. & N.A. & 89.5 & 91.2 \\
& {\OurAlg} + distillation& {\bf 85.1 / 85.3}  & {\bf 75.8} & {\bf 92.2}  &  93.2

\\ \midrule

25\%& PKD & 76.7 / 76.3 & 58.2 & 84.7 & 87.5 \\
&MixKD & 77.2 / 76.8 & 62.0 & 84.4 & 89.5 \\
&BERT-of-Theseus & 78.8 / N.A.  & 59.5  & 82.1  & 87.2  \\
& {\OurAlg} + distillation& {\bf 84.6 / 84.7}  & {\bf 72.2} & {\bf 91.4}  & {\bf 92.3} \\
\bottomrule
\end{tabular}
\end{center}

\end{table*}

\begin{table*}[htb!]

\caption{Embed {\OurAlg} into CoFi. The compression ratio is 10\%. $\text{BERT}_{\text{base}}$ indicates the fine-tuning results and is also the teacher model for knowledge distillation. CoFi and {\OurAlg} are all applied in $\text{BERT}_{\text{base}}$. The best performances are in {\bf bold}. } 
\label{tab:cofi}
\begin{center}
\begin{tabular}{l|ccccc}
\toprule
{\bf Ratio} & {\bf MNLI} & {\bf MRPC} & {\bf RTE} & {\bf QNLI} & {\bf SST-2}\\ 
~ & {Acc} & {Acc / F1} & {Acc} & {Acc} & {Acc} \\
\midrule
$\text{BERT}_{\text{base}}$ & {84.41} & {87.74 / 91.35} & {72.56} & {91.54} & {92.43}\\ \midrule
{CoFi} & {80.00}   & {84.07 / 88.50}  & {67.51} & {86.67} & {90.60} \\
{CoFi+\OurAlg} & {\bf 82.56}  & {\bf 85.54 / 89.45} & {\bf 68.23} & {\bf 89.66} & {\bf 91.51} \\
\bottomrule
\end{tabular}
\end{center}

\end{table*}

\subsection{Embed with Other Compression Method}

{\OurAlg} is a generic compression method. It can be embedded into other popular methods, such as CoFi \citep{xia2022structured}. 
CoFi is a coarse to fine-grained compression approach. It uses 3-level masks to determine which layer, heads, and neurons should be pruned. In the first level, it adds masks to MHA sub-layers and FFN sub-layers as a coarse compression. In the second level, it adds masks to the attention heads inside the MHA. In the final level, it adds masks to every neuron as the third level compression. In addition, it utilizes knowledge distillation to enhance the performance.

To embed our method into CoFi, we replace the third level masks as our method and keep the first and second level masks. Specifically, we first decompose the pre-trained weight matrices into low-rank and sparse matrices. Then, we follow the same training approach as CoFi. As for distillation, since CoFi has not released the teacher models, we download all the teacher models from Text Attack \footnote{\url{https://huggingface.co/textattack}}\citep{morris2020textattack} except teachers for the MNLI task. To obtain the MNLI teacher, we fine-tune BERT-base using following hyperparameters: learning rate is $3\times 10^{-5}$, batch size is 32, number of training epochs is 3. See Appendix \ref{sec:app_cofi} for more experiment details.

Experiment results are listed in Table \ref{tab:cofi}. We see that {\OurAlg} can improve the performance of CoFi on all datasets. For example, our method improves the accuracy by around 1\% on MRPC, RTE, and SST-2. This notable improvement shows {\OurAlg} is complementary to existing compression methods.

\section{Discussion}
\label{sec:discussion}

In Section \ref{sec:analysis}, we find Low-rank II performs much better than Low-rank I. That is, pruning out all sparse matrices is more effective than fine-tuning a low-rank matrix that is obtained from singular value thresholding. This result suggests that our method is capable of enhancing the low-rank approximation. This is because the initialization of Low-rank I is different from the pre-trained weights such that it may lose too much knowledge from the pre-trained weights. As a result, the performance drops severely on downstream tasks. Our method, on the other hand, bridges the gap between low-rank initialization and the pre-trained weight so as to retain the original knowledge stored in the pre-trained weights.
This suggests that, although the low-rank approximation alone is more efficient and concise, we should leverage the sparse approximation to guide its training process.  
Beside the improvement brought by our method, low-rank approximation, however, still have an intrinsic drawback. They ignore the diversity of neurons. Therefore, our method is crucial to remedy this drawback.

\citet{yu2017compressing, hawkins2021low} have applied the low-rank and sparse compression to CNN. They mask out some kernels in a convolution layer as the sparse approximation and add two sequential convolutional layers that are parallel to the sparse convolutional layer as the low-rank approximation. This approach, however, does not directly approximate any matrix, which makes the low-rank and sparse approximation unrelated. In addition, the kernels in CNN do not have as many dimensions as the matrices in transformer-based models. Therefore, the CNN kernels inherently have fewer ranks, thereby diminishing their efficacy when high compression rates are wanted.

We have noticed that DSEE \citep{chen2021dsee} also combines both low-rank and sparse approximation, but they apply it to the incremental matrix that is attached to a dense backbone during parameter-efficient fine-tuning. Our method, however, aims to compress the full model instead of the incremental matrix so that it is capable of saving huge memory. Moreover, DSEE masks out attention heads to realize the structure pruning while {\OurAlg} provides a more flexible and fine-grained structured pruning: it prunes neurons. Last, the low-rank approximation in {\OurAlg} is motivated by the observation that large singular values exists in large pre-trained models, while the low-rank design in DSEE is inspired by the hypothesis that the change in weights during model adaptation has a low intrinsic rank \citep{hu2022lora}.

\section{Conclusion}
\label{sec:conclustion}
We propose {\OurAlg}, a compression method for transformer models,  which combines the low-rank approximation and the structured sparse approximation. 
Experiments on natural language understanding, question answering, and natural language generation show that our method significantly surpasses previous compression approaches. Moreover, our method is particularly effective in natural language generation tasks and the setting of extremely high sparisity level. 
We show that our method is generic and complementary with other popular compression methods. Experiments show {\OurAlg} can improve the performance of CoFi and conventional iterative pruning with knowledge distillation.

\bibliography{main}

\begin{thebibliography}{53}
\providecommand{\natexlab}[1]{#1}
\providecommand{\url}[1]{\texttt{#1}}
\expandafter\ifx\csname urlstyle\endcsname\relax
  \providecommand{\doi}[1]{doi: #1}\else
  \providecommand{\doi}{doi: \begingroup \urlstyle{rm}\Url}\fi

\bibitem[Bentivogli et~al.(2009)Bentivogli, Clark, Dagan, and
  Giampiccolo]{bentivogli2009fifth}
Bentivogli, L., Clark, P., Dagan, I., and Giampiccolo, D.
\newblock The fifth pascal recognizing textual entailment challenge.
\newblock In \emph{TAC}, 2009.

\bibitem[Brown et~al.(2020)Brown, Mann, Ryder, Subbiah, Kaplan, Dhariwal,
  Neelakantan, Shyam, Sastry, Askell, et~al.]{brown2020language}
Brown, T., Mann, B., Ryder, N., Subbiah, M., Kaplan, J.~D., Dhariwal, P.,
  Neelakantan, A., Shyam, P., Sastry, G., Askell, A., et~al.
\newblock Language models are few-shot learners.
\newblock \emph{Advances in neural information processing systems},
  33:\penalty0 1877--1901, 2020.

\bibitem[Cer et~al.(2017)Cer, Diab, Agirre, Lopez-Gazpio, and
  Specia]{cer-etal-2017-semeval}
Cer, D., Diab, M., Agirre, E., Lopez-Gazpio, I., and Specia, L.
\newblock {S}em{E}val-2017 task 1: Semantic textual similarity multilingual and
  crosslingual focused evaluation.
\newblock In \emph{Proceedings of the 11th International Workshop on Semantic
  Evaluation ({S}em{E}val-2017)}, pp.\  1--14, Vancouver, Canada, August 2017.
  Association for Computational Linguistics.
\newblock \doi{10.18653/v1/S17-2001}.

\bibitem[Chen et~al.(2021)Chen, Chen, Cheng, Chen, Wang, and
  Awadallah]{chen2021dsee}
Chen, X., Chen, T., Cheng, Y., Chen, W., Wang, Z., and Awadallah, A.~H.
\newblock Dsee: Dually sparsity-embedded efficient tuning of pre-trained
  language models.
\newblock \emph{arXiv preprint arXiv:2111.00160}, 2021.

\bibitem[Dagan et~al.(2006)Dagan, Glickman, and Magnini]{BarHaim2006TheSP}
Dagan, I., Glickman, O., and Magnini, B.
\newblock The pascal recognising textual entailment challenge.
\newblock In Qui{\~{n}}onero-Candela, J., Dagan, I., Magnini, B., and
  d'Alch{\'e} Buc, F. (eds.), \emph{Machine Learning Challenges. Evaluating
  Predictive Uncertainty, Visual Object Classification, and Recognising Tectual
  Entailment}, pp.\  177--190, Berlin, Heidelberg, 2006. Springer Berlin
  Heidelberg.
\newblock ISBN 978-3-540-33428-6.

\bibitem[Dagan et~al.(2007)Dagan, Glickman, and Magnini]{Dagan2007ThePR}
Dagan, I., Glickman, O., and Magnini, B.
\newblock The pascal recognising textual entailment challenge.
\newblock In \emph{Machine Learning Challenges Workshop}, 2007.

\bibitem[Devlin et~al.(2018)Devlin, Chang, Lee, and Toutanova]{devlin2018bert}
Devlin, J., Chang, M.-W., Lee, K., and Toutanova, K.
\newblock Bert: Pre-training of deep bidirectional transformers for language
  understanding.
\newblock \emph{arXiv preprint arXiv:1810.04805}, 2018.

\bibitem[Dolan \& Brockett(2005)Dolan and
  Brockett]{dolan-brockett-2005-automatically}
Dolan, W.~B. and Brockett, C.
\newblock Automatically constructing a corpus of sentential paraphrases.
\newblock In \emph{Proceedings of the Third International Workshop on
  Paraphrasing ({IWP}2005)}, 2005.

\bibitem[Fan et~al.(2019)Fan, Grave, and Joulin]{fan2019reducing}
Fan, A., Grave, E., and Joulin, A.
\newblock Reducing transformer depth on demand with structured dropout.
\newblock \emph{arXiv preprint arXiv:1909.11556}, 2019.

\bibitem[Giampiccolo et~al.(2007)Giampiccolo, Magnini, Dagan, and
  Dolan]{giampiccolo-etal-2007-third}
Giampiccolo, D., Magnini, B., Dagan, I., and Dolan, B.
\newblock The third {PASCAL} recognizing textual entailment challenge.
\newblock In \emph{Proceedings of the {ACL}-{PASCAL} Workshop on Textual
  Entailment and Paraphrasing}, pp.\  1--9, Prague, June 2007. Association for
  Computational Linguistics.

\bibitem[Hajimolahoseini et~al.(2021)Hajimolahoseini, Rezagholizadeh,
  Partovinia, Tahaei, Awad, and Liu]{hajimolahoseinicompressing}
Hajimolahoseini, H., Rezagholizadeh, M., Partovinia, V., Tahaei, M.~S., Awad,
  O.~M., and Liu, Y.
\newblock Compressing pre-trained language models using progressive low rank
  decomposition.
\newblock 2021.

\bibitem[Han et~al.(2015)Han, Pool, Tran, and Dally]{han2015learning}
Han, S., Pool, J., Tran, J., and Dally, W.
\newblock Learning both weights and connections for efficient neural network.
\newblock \emph{Advances in neural information processing systems}, 28, 2015.

\bibitem[Hawkins et~al.(2021)Hawkins, Yang, Li, Lai, and
  Chandra]{hawkins2021low}
Hawkins, C., Yang, H., Li, M., Lai, L., and Chandra, V.
\newblock Low-rank+ sparse tensor compression for neural networks.
\newblock \emph{arXiv preprint arXiv:2111.01697}, 2021.

\bibitem[He et~al.(2020)He, Liu, Gao, and Chen]{he2020deberta}
He, P., Liu, X., Gao, J., and Chen, W.
\newblock Deberta: Decoding-enhanced bert with disentangled attention.
\newblock \emph{arXiv preprint arXiv:2006.03654}, 2020.

\bibitem[He et~al.(2021)He, Gao, and Chen]{he2021debertav3}
He, P., Gao, J., and Chen, W.
\newblock Debertav3: Improving deberta using electra-style pre-training with
  gradient-disentangled embedding sharing, 2021.

\bibitem[Hermann et~al.(2015)Hermann, Kocisky, Grefenstette, Espeholt, Kay,
  Suleyman, and Blunsom]{hermann2015teaching}
Hermann, K.~M., Kocisky, T., Grefenstette, E., Espeholt, L., Kay, W., Suleyman,
  M., and Blunsom, P.
\newblock Teaching machines to read and comprehend.
\newblock \emph{Advances in neural information processing systems}, 28, 2015.

\bibitem[Hinton et~al.(2015)Hinton, Vinyals, Dean,
  et~al.]{hinton2015distilling}
Hinton, G., Vinyals, O., Dean, J., et~al.
\newblock Distilling the knowledge in a neural network.
\newblock \emph{arXiv preprint arXiv:1503.02531}, 2\penalty0 (7), 2015.

\bibitem[Hsu et~al.(2022{\natexlab{a}})Hsu, Hua, Chang, Lou, Shen, and
  Jin]{hsu2022language}
Hsu, Y.-C., Hua, T., Chang, S., Lou, Q., Shen, Y., and Jin, H.
\newblock Language model compression with weighted low-rank factorization.
\newblock \emph{arXiv preprint arXiv:2207.00112}, 2022{\natexlab{a}}.

\bibitem[Hsu et~al.(2022{\natexlab{b}})Hsu, Hua, Chang, Lou, Shen, and
  Jin]{Hsu2022LanguageMC}
Hsu, Y.-C., Hua, T., Chang, S.-E., Lou, Q., Shen, Y., and Jin, H.
\newblock Language model compression with weighted low-rank factorization.
\newblock \emph{ArXiv}, abs/2207.00112, 2022{\natexlab{b}}.

\bibitem[Hu et~al.(2022)Hu, yelong shen, Wallis, Allen-Zhu, Li, Wang, Wang, and
  Chen]{hu2022lora}
Hu, E.~J., yelong shen, Wallis, P., Allen-Zhu, Z., Li, Y., Wang, S., Wang, L.,
  and Chen, W.
\newblock Lo{RA}: Low-rank adaptation of large language models.
\newblock In \emph{International Conference on Learning Representations}, 2022.

\bibitem[Jalali et~al.(2010)Jalali, Sanghavi, Ruan, and
  Ravikumar]{jalali2010dirty}
Jalali, A., Sanghavi, S., Ruan, C., and Ravikumar, P.
\newblock A dirty model for multi-task learning.
\newblock \emph{Advances in neural information processing systems}, 23, 2010.

\bibitem[Lagunas et~al.(2021)Lagunas, Charlaix, Sanh, and
  Rush]{lagunas2021block}
Lagunas, F., Charlaix, E., Sanh, V., and Rush, A.~M.
\newblock Block pruning for faster transformers.
\newblock \emph{arXiv preprint arXiv:2109.04838}, 2021.

\bibitem[Levesque et~al.(2012)Levesque, Davis, and
  Morgenstern]{levesque2012winograd}
Levesque, H., Davis, E., and Morgenstern, L.
\newblock The winograd schema challenge.
\newblock In \emph{Thirteenth international conference on the principles of
  knowledge representation and reasoning}, 2012.

\bibitem[Lewis et~al.(2020)Lewis, Liu, Goyal, Ghazvininejad, Mohamed, Levy,
  Stoyanov, and Zettlemoyer]{lewis-etal-2020-bart}
Lewis, M., Liu, Y., Goyal, N., Ghazvininejad, M., Mohamed, A., Levy, O.,
  Stoyanov, V., and Zettlemoyer, L.
\newblock {BART}: Denoising sequence-to-sequence pre-training for natural
  language generation, translation, and comprehension.
\newblock In \emph{Proceedings of the 58th Annual Meeting of the Association
  for Computational Linguistics}, pp.\  7871--7880, Online, July 2020.
  Association for Computational Linguistics.
\newblock \doi{10.18653/v1/2020.acl-main.703}.

\bibitem[Liang et~al.(2021)Liang, Zuo, Chen, Jiang, Liu, He, Zhao, and
  Chen]{liang2021super}
Liang, C., Zuo, S., Chen, M., Jiang, H., Liu, X., He, P., Zhao, T., and Chen,
  W.
\newblock Super tickets in pre-trained language models: From model compression
  to improving generalization.
\newblock \emph{arXiv preprint arXiv:2105.12002}, 2021.

\bibitem[Liang et~al.(2020)Liang, Hao, Shen, Zhou, Chen, Chen, and
  Carin]{Liang2020MixKDTE}
Liang, K.~J., Hao, W., Shen, D., Zhou, Y., Chen, W., Chen, C., and Carin, L.
\newblock Mixkd: Towards efficient distillation of large-scale language models.
\newblock \emph{ArXiv}, abs/2011.00593, 2020.

\bibitem[Lin(2004)]{lin-2004-rouge}
Lin, C.-Y.
\newblock {ROUGE}: A package for automatic evaluation of summaries.
\newblock In \emph{Text Summarization Branches Out}, pp.\  74--81, Barcelona,
  Spain, July 2004. Association for Computational Linguistics.

\bibitem[Liu et~al.(2019)Liu, Ott, Goyal, Du, Joshi, Chen, Levy, Lewis,
  Zettlemoyer, and Stoyanov]{liu2019roberta}
Liu, Y., Ott, M., Goyal, N., Du, J., Joshi, M., Chen, D., Levy, O., Lewis, M.,
  Zettlemoyer, L., and Stoyanov, V.
\newblock Roberta: A robustly optimized bert pretraining approach.
\newblock \emph{arXiv preprint arXiv:1907.11692}, 2019.

\bibitem[Louizos et~al.(2017)Louizos, Welling, and Kingma]{louizos2017learning}
Louizos, C., Welling, M., and Kingma, D.~P.
\newblock Learning sparse neural networks through $ l\_0 $ regularization.
\newblock \emph{arXiv preprint arXiv:1712.01312}, 2017.

\bibitem[McCarley et~al.(2019)McCarley, Chakravarti, and
  Sil]{mccarley2019structured}
McCarley, J., Chakravarti, R., and Sil, A.
\newblock Structured pruning of a bert-based question answering model.
\newblock \emph{arXiv preprint arXiv:1910.06360}, 2019.

\bibitem[Molchanov et~al.(2016)Molchanov, Tyree, Karras, Aila, and
  Kautz]{molchanov2016pruning}
Molchanov, P., Tyree, S., Karras, T., Aila, T., and Kautz, J.
\newblock Pruning convolutional neural networks for resource efficient
  inference.
\newblock \emph{arXiv preprint arXiv:1611.06440}, 2016.

\bibitem[Molchanov et~al.(2019)Molchanov, Mallya, Tyree, Frosio, and
  Kautz]{molchanov2019importance}
Molchanov, P., Mallya, A., Tyree, S., Frosio, I., and Kautz, J.
\newblock Importance estimation for neural network pruning.
\newblock In \emph{Proceedings of the IEEE/CVF Conference on Computer Vision
  and Pattern Recognition}, pp.\  11264--11272, 2019.

\bibitem[Morris et~al.(2020)Morris, Lifland, Yoo, Grigsby, Jin, and
  Qi]{morris2020textattack}
Morris, J., Lifland, E., Yoo, J.~Y., Grigsby, J., Jin, D., and Qi, Y.
\newblock Textattack: A framework for adversarial attacks, data augmentation,
  and adversarial training in nlp.
\newblock In \emph{Proceedings of the 2020 Conference on Empirical Methods in
  Natural Language Processing: System Demonstrations}, pp.\  119--126, 2020.

\bibitem[Narayan et~al.(2018)Narayan, Cohen, and Lapata]{Narayan2018xsum}
Narayan, S., Cohen, S.~B., and Lapata, M.
\newblock Don't give me the details, just the summary! topic-aware
  convolutional neural networks for extreme summarization.
\newblock \emph{ArXiv}, abs/1808.08745, 2018.

\bibitem[Noach \& Goldberg(2020)Noach and Goldberg]{Noach2020CompressingPL}
Noach, M.~B. and Goldberg, Y.
\newblock Compressing pre-trained language models by matrix decomposition.
\newblock In \emph{AACL}, 2020.

\bibitem[Paszke et~al.(2019)Paszke, Gross, Massa, Lerer, Bradbury, Chanan,
  Killeen, Lin, Gimelshein, Antiga, Desmaison, Kopf, Yang, DeVito, Raison,
  Tejani, Chilamkurthy, Steiner, Fang, Bai, and Chintala]{NEURIPS2019_9015}
Paszke, A., Gross, S., Massa, F., Lerer, A., Bradbury, J., Chanan, G., Killeen,
  T., Lin, Z., Gimelshein, N., Antiga, L., Desmaison, A., Kopf, A., Yang, E.,
  DeVito, Z., Raison, M., Tejani, A., Chilamkurthy, S., Steiner, B., Fang, L.,
  Bai, J., and Chintala, S.
\newblock Pytorch: An imperative style, high-performance deep learning library.
\newblock In \emph{Advances in Neural Information Processing Systems 32}, pp.\
  8024--8035. Curran Associates, Inc., 2019.

\bibitem[Radford et~al.(2019)Radford, Wu, Child, Luan, Amodei, Sutskever,
  et~al.]{radford2019language}
Radford, A., Wu, J., Child, R., Luan, D., Amodei, D., Sutskever, I., et~al.
\newblock Language models are unsupervised multitask learners.
\newblock \emph{OpenAI blog}, 1\penalty0 (8):\penalty0 9, 2019.

\bibitem[Rajpurkar et~al.(2016{\natexlab{a}})Rajpurkar, Zhang, Lopyrev, and
  Liang]{rajpurkar-etal-2016-squad}
Rajpurkar, P., Zhang, J., Lopyrev, K., and Liang, P.
\newblock {SQ}u{AD}: 100,000+ questions for machine comprehension of text.
\newblock In \emph{Proceedings of the 2016 Conference on Empirical Methods in
  Natural Language Processing}, pp.\  2383--2392, Austin, Texas, November
  2016{\natexlab{a}}. Association for Computational Linguistics.
\newblock \doi{10.18653/v1/D16-1264}.

\bibitem[Rajpurkar et~al.(2016{\natexlab{b}})Rajpurkar, Zhang, Lopyrev, and
  Liang]{rajpurkar2016squad}
Rajpurkar, P., Zhang, J., Lopyrev, K., and Liang, P.
\newblock Squad: 100,000+ questions for machine comprehension of text.
\newblock \emph{arXiv preprint arXiv:1606.05250}, 2016{\natexlab{b}}.

\bibitem[Romero et~al.(2014)Romero, Ballas, Kahou, Chassang, Gatta, and
  Bengio]{romero2014fitnets}
Romero, A., Ballas, N., Kahou, S.~E., Chassang, A., Gatta, C., and Bengio, Y.
\newblock Fitnets: Hints for thin deep nets.
\newblock \emph{arXiv preprint arXiv:1412.6550}, 2014.

\bibitem[Sanh et~al.(2020)Sanh, Wolf, and Rush]{sanh2020movement}
Sanh, V., Wolf, T., and Rush, A.
\newblock Movement pruning: Adaptive sparsity by fine-tuning.
\newblock \emph{Advances in Neural Information Processing Systems},
  33:\penalty0 20378--20389, 2020.

\bibitem[Socher et~al.(2013)Socher, Perelygin, Wu, Chuang, Manning, Ng, and
  Potts]{socher-etal-2013-recursive}
Socher, R., Perelygin, A., Wu, J., Chuang, J., Manning, C.~D., Ng, A., and
  Potts, C.
\newblock Recursive deep models for semantic compositionality over a sentiment
  treebank.
\newblock In \emph{Proceedings of the 2013 Conference on Empirical Methods in
  Natural Language Processing}, pp.\  1631--1642, Seattle, Washington, USA,
  October 2013. Association for Computational Linguistics.

\bibitem[Sun et~al.(2019)Sun, Cheng, Gan, and Liu]{Sun2019PatientKD}
Sun, S., Cheng, Y., Gan, Z., and Liu, J.
\newblock Patient knowledge distillation for bert model compression.
\newblock In \emph{Conference on Empirical Methods in Natural Language
  Processing}, 2019.

\bibitem[Sun et~al.(2020)Sun, Gan, Cheng, Fang, Wang, and
  Liu]{Sun2020ContrastiveDO}
Sun, S., Gan, Z., Cheng, Y., Fang, Y., Wang, S., and Liu, J.
\newblock Contrastive distillation on intermediate representations for language
  model compression.
\newblock In \emph{Conference on Empirical Methods in Natural Language
  Processing}, 2020.

\bibitem[Tahaei et~al.(2021)Tahaei, Charlaix, Nia, Ghodsi, and
  Rezagholizadeh]{tahaei2021kroneckerbert}
Tahaei, M.~S., Charlaix, E., Nia, V.~P., Ghodsi, A., and Rezagholizadeh, M.
\newblock Kroneckerbert: Learning kronecker decomposition for pre-trained
  language models via knowledge distillation.
\newblock \emph{arXiv preprint arXiv:2109.06243}, 2021.

\bibitem[Wang et~al.(2019)Wang, Singh, Michael, Hill, Levy, and
  Bowman]{wang2018glue}
Wang, A., Singh, A., Michael, J., Hill, F., Levy, O., and Bowman, S.~R.
\newblock {GLUE}: A multi-task benchmark and analysis platform for natural
  language understanding.
\newblock In \emph{International Conference on Learning Representations}, 2019.

\bibitem[Warstadt et~al.(2019)Warstadt, Singh, and
  Bowman]{warstadt-etal-2019-neural}
Warstadt, A., Singh, A., and Bowman, S.~R.
\newblock Neural network acceptability judgments.
\newblock \emph{Transactions of the Association for Computational Linguistics},
  7:\penalty0 625--641, 2019.
\newblock \doi{10.1162/tacl_a_00290}.

\bibitem[Williams et~al.(2018)Williams, Nangia, and
  Bowman]{williams-etal-2018-broad}
Williams, A., Nangia, N., and Bowman, S.
\newblock A broad-coverage challenge corpus for sentence understanding through
  inference.
\newblock In \emph{Proceedings of the 2018 Conference of the North {A}merican
  Chapter of the Association for Computational Linguistics: Human Language
  Technologies, Volume 1 (Long Papers)}, pp.\  1112--1122, New Orleans,
  Louisiana, June 2018. Association for Computational Linguistics.
\newblock \doi{10.18653/v1/N18-1101}.

\bibitem[Xia et~al.(2022)Xia, Zhong, and Chen]{xia2022structured}
Xia, M., Zhong, Z., and Chen, D.
\newblock Structured pruning learns compact and accurate models.
\newblock \emph{arXiv preprint arXiv:2204.00408}, 2022.

\bibitem[Xu et~al.(2020)Xu, Zhou, Ge, Wei, and Zhou]{Xu2020BERTofTheseusCB}
Xu, C., Zhou, W., Ge, T., Wei, F., and Zhou, M.
\newblock Bert-of-theseus: Compressing bert by progressive module replacing.
\newblock In \emph{Conference on Empirical Methods in Natural Language
  Processing}, 2020.

\bibitem[Yu et~al.(2017)Yu, Liu, Wang, and Tao]{yu2017compressing}
Yu, X., Liu, T., Wang, X., and Tao, D.
\newblock On compressing deep models by low rank and sparse decomposition.
\newblock In \emph{Proceedings of the IEEE conference on computer vision and
  pattern recognition}, pp.\  7370--7379, 2017.

\bibitem[Zhang et~al.(2022)Zhang, Zuo, Liang, Bukharin, He, Chen, and
  Zhao]{zhang2022platon}
Zhang, Q., Zuo, S., Liang, C., Bukharin, A., He, P., Chen, W., and Zhao, T.
\newblock Platon: Pruning large transformer models with upper confidence bound
  of weight importance.
\newblock In \emph{International Conference on Machine Learning}, pp.\
  26809--26823. PMLR, 2022.

\bibitem[Zhu \& Gupta(2017)Zhu and Gupta]{zhu2017prune}
Zhu, M. and Gupta, S.
\newblock To prune, or not to prune: exploring the efficacy of pruning for
  model compression.
\newblock \emph{arXiv preprint arXiv:1710.01878}, 2017.

\end{thebibliography}
\bibliographystyle{icml2023}

\newpage
\appendix
\onecolumn
\section{GLUE Dataset Statistics}
\label{app:dataset-glue}

We present the dataset statistics of GLUE \cite{wang2018glue} in the following table. 
\begin{table*}[htb!]
	\begin{center}
		\begin{tabular}{l|l|c|c|c|c|c}
			\toprule 
			\bf Corpus &Task& \#Train & \#Dev & \#Test   & \#Label &Metrics\\ \midrule
			\multicolumn{6}{@{\hskip1pt}r@{\hskip1pt}}{Single-Sentence Classification (GLUE)} \\ \hline
			CoLA & Acceptability&8.5k & 1k & 1k & 2 & Matthews corr\\ \hline
			SST & Sentiment&67k & 872 & 1.8k & 2 & Accuracy\\ \midrule
			\multicolumn{6}{@{\hskip1pt}r@{\hskip1pt}}{Pairwise Text Classification (GLUE)} \\ \hline
			MNLI & NLI& 393k& 20k & 20k& 3 & Accuracy\\ \hline
			RTE & NLI &2.5k & 276 & 3k & 2 & Accuracy \\ \hline
			QQP & Paraphrase&364k & 40k & 391k& 2 & Accuracy/F1\\ \hline
			MRPC & Paraphrase &3.7k & 408 & 1.7k& 2&Accuracy/F1\\ \hline
			QNLI & QA/NLI& 108k &5.7k&5.7k&2& Accuracy\\ \midrule
			\multicolumn{5}{@{\hskip1pt}r@{\hskip1pt}}{Text Similarity (GLUE)} \\ \hline
			STS-B & Similarity &7k &1.5k& 1.4k &1 & Pearson/Spearman corr\\ \bottomrule
		\end{tabular}
	\end{center}
	\vskip -0.05in
	\caption{Summary of the GLUE benchmark.}
	\label{tab:glue}
\end{table*}

\section{Natural Language Understanding}
\label{sec:app_nlu}
\subsection{Training Details} 

{\bf Implementation Details.} The implementation of {\OurAlg} is based on publicly available Huggingface \citep{NEURIPS2019_9015} code-base \footnote{\url{https://github.com/huggingface/transformers/tree/main/examples/pytorch}}. 

{\bf Hyper-parameter Details.} 

We select the proportion of the parameters $r$ of all low-rank matrices over all pre-trained parameters from $\{1\%, 2\%, 3\%, 5\%\}$ and present the best final ratio we choose as below. Neuron importance scores are often unstable during training due to the variance between different data from different batches and different training dynamics between iterations (e.g. dropout) \citep{zhang2022platon}. In addition to involving exponential moving of average in calculating the neuron importance score, we attempt large batch sizes to calculate a more smooth and accurate importance score. We find out large batch sizes are profoundly helpful in most GLUE tasks.Therefore, we apply a large batch size on most tasks in GLUE.

For the choice of pruning hyperparameters, we follow the pruning schedule of \citet{zhang2022platon},i.e. the training epochs, initial warm up, and final warm up. We only change the warm up steps to accommodate the change in batch sizes as the total training steps will change when batch size changes. We also use the same $\beta$ as \citet{zhang2022platon} except for some minor changes in CoLA and RTE task.

Table \ref{tab:app_glue_setup} summarizes the detailed hyperparameters for each task used in pruning DeBERTaV3-base. Table \ref{tab:app_glue_setup_bert} summarizes the detailed hyperparameters for each task used in pruning BERT-base.

\begin{table*}[h!]
\vspace{-1mm}
\caption{Hyper-parameter setup of {\OurAlg} for GLUE benchmark for pruning DeBERTaV3-base.}
\label{tab:app_glue_setup}
\begin{center}
\begin{small}
\begin{tabular}{l|c|cccccccc}
\toprule
\multirow{1}*{\bf Ratio} & {\bf Hyper-parameter} & {\bf MNLI} & {\bf RTE} & {\bf QNLI}  & {\bf MRPC} & {\bf QQP } & {\bf SST-2} & {\bf CoLA} & {\bf STS-B} \\ 
\midrule 
~ & \# epochs &   8  &  20 & 10 & 10 & 10 & 6 & 15 & 15 \\
~ & Batch size &   256 &  128 & 256 & 64 & 256 & 256 & 256 & 16 \\
~ & Learning rate & $ 9\times 10^{-5}$ & $1\times 10^{-4}$ & $ 5\times 10^{-5} $ & $ 1\times 10^{-4} $  & $ 5\times 10^{-5} $ & $ 8\times 10^{-5}$ & $ 3\times 10^{-4} $ & $1\times 10^{-4}$ \\
~ & $ t_i $  & 675 & 25 & 250 & 38 & 675 & 125 & 62 & 500 \\
~ & $ t_f $ &  3375 & 150 & 1500 & 112 & 2750 & 1250 & 187 & 2500 \\
\midrule
\multirow{2}*{20\%}
~ & $ \beta $ & 0.85 & 0.75 & 0.85 & 0.85 & 0.85 & 0.85 & 0.7 & 0.85 \\
~ & $ r $ &  5\% & 2\% & 3\% & 5\% & 2\% & 1\% & 2\% & 2\%  \\
\midrule 
\multirow{2}*{15\%} 
~ & $ \beta $ &  0.85 & 0.85 & 0.85 & 0.85 & 0.85 & 0.85 & 0.75 & 0.85 \\
~ & $ r $ &  3\% & 2\% & 1\% &1\% & 3\% & 3\% & 2\% & 2\% \\
\midrule 
\multirow{2}*{10\%} 
~ & $ \beta $ & 0.85 & 0.85 & 0.85 & 0.85 & 0.85 & 0.85 & 0.8 & 0.85 \\
~ & $ r $ & 3\% & 2\% & 5\% & 3\% & 5\% & 1\% & 2\% & 2\% \\
\bottomrule
\end{tabular}
\end{small}
\end{center}
\end{table*}

\begin{table*}[h!]
\vspace{-1mm}
\caption{Hyper-parameter setup of {\OurAlg} for GLUE benchmark for pruning BERT-base.}
\label{tab:app_glue_setup_bert}
\begin{center}
\begin{small}
\begin{tabular}{l|c|cccccccc}
\toprule
\multirow{1}*{\bf Ratio} & {\bf Hyper-parameter} & {\bf MNLI} & {\bf RTE} & {\bf QNLI}  \\ 
\midrule 
~ & \# epochs &   8  &  20 & 10  \\
~ & Batch size &   256 &  128 & 256 \\
~ & Learning rate & $ 5\times 10^{-5}$ & $5\times 10^{-5}$ & $ 5\times 10^{-5} $ \\
~ & $ t_i $  & 675 & 25 & 250  \\
~ & $ t_f $ &  3375 & 150 & 1500  \\
\midrule
\multirow{2}*{20\%}
~ & $ \beta $ & 0.85 & 0.60 & 0.85  \\
~ & $ r $ &  5\% & 2\% & 2\%  \\
\midrule 
\multirow{2}*{15\%} 
~ & $ \beta $ &  0.85 & 0.7 & 0.85  \\
~ & $ r $ &  5\% & 2\% & 3\% \\
\midrule 
\multirow{2}*{10\%} 
~ & $ \beta $ & 0.85 & 0.50 & 0.85 \\
~ & $ r $ & 5\% & 2\% & 3\%  \\
\bottomrule
\end{tabular}
\end{small}
\end{center}
\end{table*}

\section{Question Answering}
\label{sec:app_squad}

\subsection{Dataset}
Following \citet{sanh2020movement}, we also choose SQuAD v1.1 \cite{rajpurkar2016squad} to evaluate the performance of {\OurAlg} on question answering task.

\subsection{Training Details}
We set the batch size as 16, the number of epochs for fine-tuning as 10, the optimizer as AdamW and the learning rate as $ 5\times 10^{-5} $ for all experiments. Similarly, we follow the pruning schedule of \citet{zhang2022platon},i.e. we take the same initial warm up steps and final warm up steps. We use the same settings for all sparsities. The hyperparameters are summarized specifically in Table \ref{tab:app_squad_setup}. We use the hyperparameters in Table \ref{tab:app_squad_setup} for pruning both DeBERTaV3-base and BERT-base.

\begin{table*}[htb!]
\vspace{-2mm}
\caption{Hyper-parameter setup of {\OurAlg} on question answering tasks (SQuAD v1.1, \citet{rajpurkar2016squad}).}
\vspace{3mm}
\label{tab:app_squad_setup}
\begin{center}
\begin{small}
\begin{tabular}{l|cccccccc}
\toprule
\multirow{1}*{\bf Task} & \# epochs & Batch size  & Learning rate & $ t_i $ & $ t_f $ & $r$ & $\beta$ \\ 
\midrule 
SQuAD  & 10&16& $ 5\times 10^{-5} $ &5400&22000&5\% & 0.85 \\
\bottomrule
\end{tabular}
\end{small}
\end{center}
\end{table*}

\section{Natural Language Generation}
\label{sec:app_nlg}
\subsection{Training Details}
We set the batch size as 32, the number of training epoch as 10. We choose Adam as the optimizer and try learning rate from $\{6\times10^{-6},1\times10^{-5}, 2\times10^{-5}, 4\times10^{-5}, 6\times10^{-5}, 1\times10^{-4}\}$. We find the optimal learning rate is $4\times10^{-5}$. We also adjust the sparse approximation ratio, choosing from {5\%, 10\%, 15\%, 20\%}. We find the best sparse ratio is 10\%. We also fix the initial warm up steps $t_i$ as 12800, final warm up steps $t_f$ as 51200, and $\beta$ as 0.85.

\section{Combination with Knowledge Distillation}
\label{sec:app_KD}
\subsection{Teacher Models}
For DeBERTaV3-base teacher models, We trained the teacher models following the hyperparameters of in \citet{he2021debertav3}'s official repository \footnote{\url{https://github.com/microsoft/DeBERTa}}.The performance of the teacher model are shown in Table \ref{tab:glue_distillation}.

For BERT-base teacher models, we used the teacher models released on Huggingface by textattack \footnote{\url{https://huggingface.co/textattack}}.

\subsection{Training Details}
We first prune the model following the details Table \ref{tab:app_glue_setup}. Then, we conduct layerwise distillation with distillation coefficient $\alpha$ as 15. The other training hyperparameters are listed as below.

\begin{table*}[htb!]
\vspace{-2mm}
\caption{Hyper-parameter setup of {\OurAlg} on knowledge distillation with DeBERTaV3-base.}
\vspace{3mm}
\label{tab:app_kd_setup}
\begin{center}
\begin{small}
\begin{tabular}{l|cccccc}
\toprule
\multirow{1}*{\bf Task} & \# epochs & Batch size  & Learning rate & alpha\_output & alpha\_layer \\ 
\midrule 
MNLI  & 50&32& $ 9\times 10^{-5} $ & 0 & 15 \\
\midrule 
SQuAD  & 50&16& $ 5\times 10^{-5} $ & 0 & 15 \\
\midrule 
SST-2  & 50&32& $ 8\times 10^{-5} $ & 0 & 15 \\
\midrule 
RTE  & 50&16& $ 1\times 10^{-4} $ & 0 & 15 \\
\bottomrule
\end{tabular}
\end{small}
\end{center}
\end{table*}

\begin{table*}[htb!]
\vspace{-2mm}
\caption{Hyper-parameter setup of {\OurAlg} on knowledge distillation with BERT-base.}
\vspace{3mm}
\label{tab:app_kd_setup_bert}
\begin{center}
\begin{small}
\begin{tabular}{l|cccccc}
\toprule
\multirow{1}*{\bf Task} & \# epochs & Batch size  & Learning rate & alpha\_output & alpha\_layer \\ 
\midrule 
MNLI  & 50&32& $ 9\times 10^{-5} $ & 0 & 15 \\
\midrule 
RTE & 50&16& $ 5\times 10^{-5} $ & 0 & 15 \\
\midrule 
QNLI  & 50&32& $ 5\times 10^{-5} $ & 5 & 5 \\
\midrule 
SST-2  & 50&32& $ 3\times 10^{-4} $ & 1 & 1 \\
\bottomrule
\end{tabular}
\end{small}
\end{center}
\end{table*}

\section{Combination with CoFi}
\label{sec:app_cofi}
\subsection{Teacher Models}
As CoFi has not released the teacher models, we download all the teacher models from Text Attack \footnote{\url{https://huggingface.co/textattack}}\citep{morris2020textattack} except teachers for the MNLI task. To obtain the MNLI teacher, we fine-tune BERT-base using following hyperparameters: learning rate: $3\times 10^{-5}$, batch size: 32, training epochs: 3.

\subsection{Training Details}
CoFi masks out hidden states to control the remaining parameters while our method compress matrices directly, so the total compression ratio is easily calculated as $\rm{ratio} = \rm{ratio}_{\rm{CoFi}} \times \rm{ratio}_{\rm{\OurAlg}}$. We choose $\rm{ratio}_{\rm{\OurAlg}}=0.5$ and $\rm{ratio}_{\rm{CoFi}}=0.2$ for 10\% total compression ratio.

For {\OurAlg} part, we use the same hyperparameters in Appendix \ref{sec:app_nlu}. As for CoFi, please refer the training schedule and the rest hyperparameters to its official repository \footnote{\url{https://github.com/princeton-nlp/CoFiPruning}}.


\end{document}